\documentclass{article}

\usepackage{microtype}
\usepackage{graphicx}
\usepackage{subcaption}
\usepackage{booktabs} 

\usepackage{xcolor}

\usepackage{hyperref}

\usepackage[preprint]{icml2026}

\usepackage{amsmath}
\usepackage{amssymb}
\usepackage{mathtools}
\usepackage{amsthm}

\usepackage[utf8]{inputenc} 
\usepackage[T1]{fontenc}    
\usepackage{url}            
\usepackage{amsfonts}       
\usepackage{nicefrac}       
\usepackage{xcolor}         
\usepackage{multirow}
\usepackage{pifont}
\usepackage{sidecap}
\usepackage{comment}
\usepackage{makecell}
\usepackage{enumitem} 
\usepackage{tabularx}
\usepackage{placeins}

\usepackage{import}   
\usepackage{transparent} 

\usepackage[capitalize,noabbrev]{cleveref}

\theoremstyle{plain}

\theoremstyle{definition}

\theoremstyle{remark}

\usepackage[textsize=tiny]{todonotes}

\icmltitlerunning{SiLIF: Structured State Space Model Dynamics and Parametrization for Spiking Neural Networks}

\begin{document}

\twocolumn[
  \icmltitle{SiLIF: Structured State Space Model Dynamics and  \\
    Parametrization for Spiking Neural Networks}

  \icmlsetsymbol{equal}{*}

    \begin{icmlauthorlist}
      \icmlauthor{Maxime Fabre}{equal,fzj,rug}
      \icmlauthor{Lyubov Dudchenko}{equal,eth}
      \icmlauthor{Younes Bouhadjar}{fzj}
      \icmlauthor{Emre Neftci}{fzj,rwth}
    \end{icmlauthorlist}
    
    \icmlaffiliation{fzj}{Peter Grünberg Institute, Forschungszentrum Jülich, Jülich, Germany}
    \icmlaffiliation{rug}{Groningen Cognitive Systems and Materials Center (CogniGron), University of Groningen, Groningen, The Netherlands}
    \icmlaffiliation{eth}{ETH Zürich, Zürich, Switzerland}
    \icmlaffiliation{rwth}{RWTH Aachen University, Aachen, Germany}
    
    \icmlcorrespondingauthor{Maxime Fabre}{m.fabre@fz-juelich.de}
    \icmlcorrespondingauthor{Lyubov Dudchenko}{ldudchenko@ethz.ch}

  \icmlkeywords{Machine Learning, ICML}

  \vskip 0.3in
]

\printAffiliationsAndNotice{} 
\begin{abstract}
Multi-state spiking neurons combine sparse binary activations with rich second-order nonlinear recurrent dynamics, making them a promising alternative to standard deep learning models.
However, gradient propagation through these dynamics often leads to instabilities that hinder scalability and performance.
Inspired by the stable training and strong performance of state space models (SSMs) on long sequences, we introduce two SSM-inspired Leaky Integrate-and-Fire (SiLIF) neuron models. The first extends a two-state neuron with a learnable discretization timestep and logarithmic reparametrization, while the second additionally incorporates the initialization scheme and structure of complex-state SSMs, enabling oscillatory regimes. 
Our two SiLIF models achieve new state-of-the-art performance among spiking neuron models on both event-based and raw-audio speech recognition datasets. 
We further demonstrate a favorable performance-efficiency trade-off compared to SSMs, even surpassing them while using half the computational cost through the use of synaptic delays. Our code is available at \href{https://github.com/Maxtimer97/SSM-inspired-LIF}{https://github.com/Maxtimer97/SSM-inspired-LIF}.
\end{abstract}

\begin{figure*}[t]
\begin{center}
    \includegraphics[width=6in]{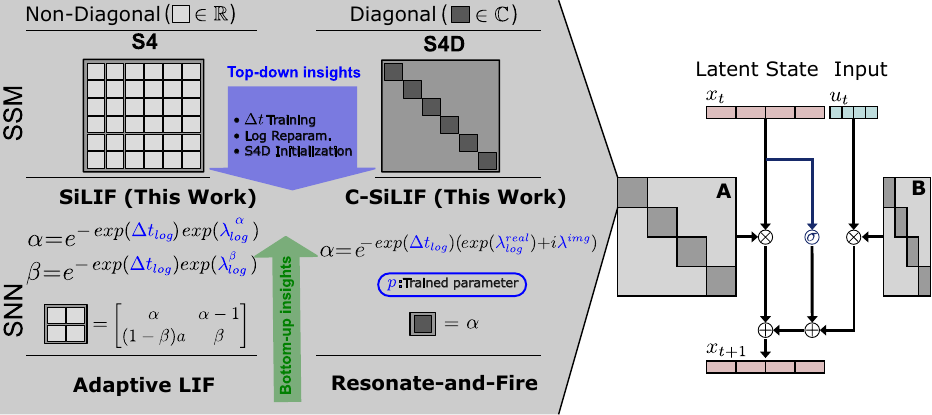}
\end{center}
\caption{Our proposed method distills features from modern structured state space models (SSMs) to build novel high-performing spiking neural networks (SNNs). The SiLIF model is an enhancement of the adaptive LIF (AdLIF) neuron, achieved by reparametrizing its state transition matrix according to the S4 model. The C-SiLIF additionally exploits the complex representation and specific initialization of the S4D model, drawing a parallel with the resonate-and-fire (RF) neuron. The right part of the figure illustrates the similar structure of SSMs and SNNs, plus the SNN-exclusive reset mechanism.}
\label{fig:mainScheme}
\end{figure*}

\section{Introduction}
Spiking Neural Networks (SNNs) and their realization in neuromorphic hardware are gaining interest as promising low-power alternatives to artificial neural networks (ANNs) \cite{neurobench, Lyes}.
In addition to being compact recurrent units with strong temporal encoding capabilities, the biologically inspired binary spiking and the nonlinear feedback mechanism, known as reset, enable sparse event-based communication.\footnote{In the following, we refer to such models as reset-based spiking models, in contrast to non-reset models that only offer linear recurrence and non-spiking models that do not employ binary or at least graded spiking.}
Over the past decade, methods leveraging backpropagation through time (BPTT) to train SNNs similarly to recurrent neural networks (RNNs) have significantly improved SNN performance on a variety of benchmarks, thanks to the use of surrogate gradients \cite{kaiser2020synaptic, zenke2018superspike}. 
Nonetheless, due to vanishing and exploding gradients and saturation caused by binary spikes, the training dynamics of SNNs can remain unstable, limiting their performance \citep{herranz2022stabilizing}.

A potential avenue for addressing these challenges is through the perspective of state space models (SSMs) \cite{s4, S4D, orvieto2023resurrecting}. 
SSMs have recently demonstrated superior long-sequence modeling performance on various benchmarks, often outperforming traditional RNNs.
This performance gain is believed to stem from SSMs' linear recurrent dynamics (unlike traditional RNNs or SNNs) along with several training optimizations \cite{orvieto2023resurrecting}.
In fact, dedicated parametrizations and initialization introduced with deep learning SSMs such as the S4 model \citep{s4} have been shown to provide robust and backpropagation-friendly recurrent state dynamics \citep{zucchet2024}.\\

This study leverages the strengths of the S4 model to enhance nonlinear recurrent SNNs on spatiotemporal sequences. 
Specifically, we establish parallels between the Adaptive Leaky Integrate-and-Fire (AdLIF) neuron \cite{Bittar2022}, the Resonate and Fire (RF) neuron \cite{Izhikevich2001}, and SSMs. 
Building on these parallels, we propose two models: (1) The SSM-Inspired LIF (SiLIF) model, a S4-based enhancement of the AdLIF neuron with timestep training and logarithmic reparametrization. (2) The complex-valued SSM-Inspired LIF (C-SiLIF) model, a novel top-down designed spiking neuron employing the diagonal S4 (S4D) \cite{S4D} resonant dynamics, parametrization, and initialization with a matching reset mechanism. 

Following previous work, we evaluate these reset-based spiking models on a suite of keyword classification datasets spanning event-based and raw-audio domains. 
Thanks to their integrated SSM-inspired features and improved training stability, our SiLIF models achieve new state-of-the-art (SOTA) performance across all covered tasks and even outperform SSM models with synaptic delays. Our key contributions are:
\begin{enumerate}[nosep]
    \item Establishing theoretical bridges between SSMs and spiking neural networks (SNNs), revealing key insights into their shared and specific training dynamics.
    \item Proposing two novel reset-based spiking neuron models that incorporate SSM-derived dynamics and parametrizations.
    \item Demonstrating significant accuracy improvements on audio classification tasks while maintaining efficiency, surpassing previous reset-based models.
    \item Demonstrating Pareto-optimal accuracy-efficiency balance compared to SSM models across speech recognition tasks, and surpassing them with half the number of synaptic operations (SOPs) - our proxy for MACs in SNN weight matrices - through the insertion of synaptic delays.
\end{enumerate}

\section{Related Work}
\label{rel_work}
\subsection{Spiking Neural Networks for Spoken Word Classification Tasks}
Training SNNs has been vastly improved by techniques from deep learning, such as the surrogate gradient method \cite{Neftci19}, error backpropagation mechanisms \cite{Lee16}, and ANN-to-SNN conversion \cite{Bu2022}.
One promising application for spiking neurons is sequence processing, such as speech recognition.\\
Inspired by previous neuroscience studies \cite{Brette2005, gerstner2014neuronal, Salaj2021,Yin2021}, \citet{Bittar2022} successfully introduced an Adaptive Leaky Integrate-and-Fire (AdLIF) neuron for audio classification.
Several works subsequently refined the AdLIF model, either by constraining parameters during training \citep{co-learning-delays}, or by employing different discretization approaches \citep{baronig2025}.
However, these only led to marginal improvements on key benchmarks. Our work significantly advances performance on benchmarks by extending the AdLIF model through SSM-inspired training dynamics.

In parallel, the resonate-and-fire (RF) neuron model~\cite{Izhikevich2001} was also extended in deep SNN works~\citep{alkhamissi2021, Frady2022, higuchi2024}, but their performance remains below the state-of-the-art (SOTA), even when scaled up and employing an SSM structure \citet{huber2025}. Our improved initialization and parametrization, inspired by SSMs, lead to the first RF-style spiking neuron achieving SOTA performance on audio classification benchmarks. \citet{karilanova2025zero} also draw parallels between SSMs and SNNs but do not derive enhancements for neuron models.

Synaptic delays further enhance SNNs' temporal processing capabilities, achieving SOTA accuracy on event-based datasets using LIF neurons~\citep{hammouamri2024}. 
Synaptic delays can also be combined with richer models. For instance, \citet{co-learning-delays} extend the AdLIF model with learnable delays to improve accuracy on keyword spotting tasks.
We note that delay-based models require per-synapse buffers to store incoming spikes and deliver them when the delay has passed, thus requiring additional on-chip memory, registers, and compute.
While our models meet or exceed the accuracy of several delay-based models, we show that adding synaptic delays to the SiLIF achieves new SOTA performance.

\subsection{State Space Models for Spiking and Event-driven Deep Neural Networks}
Linear recurrent SSMs emerged as reference models for long sequences \cite{s4, Hippo, S5, orvieto2023resurrecting}, outperforming vanilla transformer models in the long range arena benchmark \citep{tay2020long}.
Gated SSM variants, such as Mamba \cite{gu2024mamba}, further improve SSM performance with selective data-dependent gating \cite{gu2024mamba, dao2024transformers, beck2024xlstm}. Such gating mechanisms endow SSMs with LSTM-style input and forget gates, while maintaining their linear recurrency and fast, parallelized training. While these models excel on long sequence tasks, our SiLIF models exhibit a more favorable accuracy-efficiency balance on shorter-sequence speech tasks.

Several works propose to attach neuron dynamics and binary spiking to SSMs to enhance their efficiency \citep{huang2024prfparallelresonateneuron, s6, shen2025spikingssms}. 
Notably, SpikingSSM \citep{shen2025spikingssms} appends a LIF neuron to the S4D model \citep{S4D} and proposes a surrogate network to maintain fast training. The modifications, however, incur accuracy degradation that is not compensated by the efficiency gains, leaving our SiLIF models Pareto-optimal in comparison. 

In parallel, conventional deep SSMs were applied to event-based data \cite{schone2024, statespacemodelsevent}. Event-SSM \citep{schone2024} reaches, for instance, new SOTA performance on event-based keyword classification but at a significant computational overhead, resulting in unfavorable accuracy-efficiency trade-offs compared to our SiLIF models.

\section{Methods}
\label{methods}
\subsection{State Space Models}
\label{subsection:StructuredStateSpaces}
State space models are based on the linear latent state dynamics (Eq.~\ref{eq:SSMeq} below). Each scalar input feature $u \in \mathbb{R}$ is expanded to $N$ dimensions through a complex vector $ \mathbf{B} \in \mathbb{C}^{N}$, yielding the complex state $x \in \mathbb{C}^{N}$, which is recursively updated through the transition matrix $\mathbf{A} \in \mathbb{C}^{N \times N}$. The readout $y \in \mathbb{R}$ is defined as the real part $\mathrm{Re}(\cdot)$ of the state projection through $\mathbf{C} \in \mathbb{C}^{N}$ plus the direct signal projection through $\mathbf{D} \in \mathbb{R}$. The discretized form for a timestep $\Delta t$ is obtained with the zero-order hold (ZOH) method:
\begin{equation}
\begin{aligned}
    x_t &= \bar{\mathbf{A}} x_{t-1} + \bar{\mathbf{B}} u_t\\
    y_t &= \mathrm{Re}(\bar{\mathbf{C}} x_t) + \bar{\mathbf{D}} u_t
\end{aligned}
\label{eq:SSMeq}
\end{equation}
where
$\bar{\mathbf{A}} = e^{\mathbf{A}\Delta t}$, $\quad \bar{\mathbf{B}} = \mathbf{A}^{-1} \left( \bar{\mathbf{A}} - \mathbf{I} \right) \mathbf{B}$, $\quad \bar{\mathbf{C}} = \mathbf{C}, \quad \bar{\mathbf{D}} = \mathbf{D}$. These dynamics generalize to an input of dimension $H$ by instantiating separate SSM blocks for each input dimension, resulting in block-diagonal matrices $\mathbf{A}$ and $\mathbf{B}$ (Fig.~\ref{fig:mainScheme}). 
Following \citet{s4}, SSM blocks are then typically placed in successive layers interleaved with multi-layer perceptrons (MLP) and gated linear units (GLU). 
\subsection{AdLIF Neurons in the SSM Domain}
Second-order spiking neurons take inspiration from the dynamics of biological neurons to enable complex and scalable temporal processing \cite{Izhikevich2001, Brette2005, gerstner2014neuronal, Salaj2021, Yin2021}. 
Our work builds on the Adaptive Leaky Integrate-and-Fire (AdLIF) model \cite{Bittar2022}, which achieved competitive performance on speech recognition tasks.  
The AdLIF neuron follows second-order dynamics, where an adaptation variable $w$ is coupled with the neuron's membrane potential $u$:
\begin{equation}
\begin{aligned}
u_t & = \alpha u_{t-1} + (1 - \alpha)(I_t - w_{t-1}) - \theta s_{t-1} \\
w_t & = \beta w_{t-1} + a u_{t-1} + b s_{t-1} \\
s_t & = (u_t \geq \theta)
\end{aligned}
\label{eq:adlif}
\end{equation}
where $t$ is the timestep, $I_t$ is the input signal, $s_t$ the potential binary spike, $\theta$ is the spiking threshold, and $a$ and $b$ sub- and above-threshold feedback parameters. 
The decay constants $\alpha$ and $\beta$ are defined as $\alpha = e^{-\Delta t/\tau_u}$ and $\beta = e^{-\Delta t/\tau_w}$, where $\Delta t$ is the discretization timestep and $\tau_u$ and $\tau_w$ are the continuous-space time constants of the $u$ and $w$ variables. 
The model is trained in the discrete domain, using the decay constants $\alpha$ and $\beta$ as learnable parameters, without involving $\tau_u$ or $\tau_w$. 

In its subthreshold regime, the AdLIF neuron (Eq.~\ref{eq:adlif}) can be mapped to a linear SSM block (Eq.~\ref{eq:SSMeq}) identifying the state $x_t=\begin{bmatrix}u_t\\w_t\end{bmatrix}$ and using parameters: 
\begin{equation}
\mathbf{\bar{A}}=\begin{bmatrix}\alpha&\alpha-1\\a&\beta\end{bmatrix},
\mathbf{\bar{B}}=\begin{bmatrix}1-\alpha\\0\end{bmatrix}, 
\mathbf{\bar{C}}=\begin{bmatrix}1\\0\end{bmatrix}, \mathbf{\bar{D}}=0
\label{eq:adlif_ssm}
\end{equation}
and noting $s_t=0$ in the subthreshold regime.
The only remaining differences between AdLIF and SSMs are the SSM's state expansion and the spike-triggered feedback. 
This feedback captures the neuron's reset mechanism, which is applied to the states $u$ and $w$ through $\theta$ and $b$, respectively, making the state dynamics nonlinearly dependent on the neuron membrane potential.
This feature plays a crucial role in the sparsity and nonlinear adaptation of SNNs. 
In the following sections, we present two model variants that merge efficient SSM features with nonlinear adaptive spiking neuron models. 

We note that bridges between SNNs and SSMs have already been investigated in previous work \citep{huber2025, karilanova2025zero}, but without leveraging the associated training dynamics. 
We study these discrepancies in more detail in the next two paragraphs.

\subsection{SiLIF Model: SSM-inspired Parametrization}
The SSM-inspired LIF neuron (SiLIF) is derived by integrating SSM features into the AdLIF neuron. 
Firstly, we note that structured SSMs such as S4 \citep{s4} are generally used in their discretized form (Eq.~\ref{eq:SSMeq}), while employing continuous-domain state-space variables $\mathbf{A}$ and $\mathbf{B}$ as trained parameters.
Following this approach, rather than directly training the discrete decay factors $\alpha$ and $\beta$ of the AdLIF (Eq.~\ref{eq:adlif}), we train their continuous-domain counterparts $\lambda^{\alpha} = \tau_u^{-1}$ and $\lambda^{\beta} = \tau_w^{-1}$. 

Unlike most prior SNN studies, which treat the timestep $\Delta t$ as a fixed dataset-dependent constant, following S4's approach \cite{s4}, we define it as a learnable parameter for each SiLIF neuron.
This additional flexibility allows the model to cover a wider range of transition dynamics regimes across neurons (Fig.~\ref{fig:EVs}).

Additionally, we implement S4's logarithmic reparametrization to enhance numerical stability and improve optimization smoothness. 
We train the logarithmic counterparts $\lambda^{\alpha}_{log}$, $\lambda^{\beta}_{log}$ and $\Delta t_{log}$, such that $\lambda^{\alpha} = \exp(\lambda^{\alpha}_{log}), \, \lambda^{\beta} = \exp(\lambda^{\beta}_{log})$ and $\Delta t = \exp(\Delta t_{log})$. 
This transformation reduces the risk of vanishing and exploding gradients, thus improving the training stability of recurrent architectures \citep{zucchet2024}.  
We extend and evaluate this observation on SNNs for the first time here. 

Furthermore, since $\exp(\lambda^{\alpha}_{log})$, $\exp(\lambda^{\beta}_{log})$ and $\exp(\Delta t_{log})$ are strictly positive, the effective decay factors are constrained within a stable regime $0\leq \alpha , \beta \leq 1$ without requiring explicit clamping, as previously employed in AdLIF models for stability \citep{Bittar2022}.
The resulting SiLIF model 
is described in Alg. \ref{alg:siLIF_alg}. 
\begin{algorithm}[htbp!]
\footnotesize
\caption{SiLIF: an SSM-inspired LIF neuron}\label{alg:siLIF_alg}
\begin{algorithmic}

\STATE \textbf{Initialize trainable parameters:} \\
\STATE $\lambda^{\alpha}_{log} \sim \mathcal{U}(\log(\lambda^{\alpha}_{\text{min}}),\log(\lambda^{\alpha}_{\text{max}}))$ , \\ 
\STATE $\lambda^{\beta}_{log} \sim \mathcal{U}(\log(\lambda^{\beta}_{\text{min}}), \log(\lambda^{\beta}_{\text{max}}))$, $\Delta t_{log}  \leftarrow \log(\Delta t_0)$, \\ 
\STATE $a \sim \mathcal{U}(a_{\text{min}}, a_{\text{max}})$, $b \sim \mathcal{U}(b_{\text{min}}, b_{\text{max}})$\\ 
\STATE \textbf{Initialize states:} \\$u \sim \mathcal{U}(0, 1)$, $w \sim \mathcal{U}(0, 1)$, $s \sim \mathcal{U}(0, 1)$ 
\STATE\vspace{-.7em}
\STATE \textbf{Forward pass:} 
\STATE $\alpha \leftarrow \exp(-\exp(\lambda^{\alpha}_{log}) \cdot \exp(\Delta t_{log}))$, \\  $\beta \leftarrow \exp(-\exp(\lambda^{\beta}_{log}) \cdot \exp(\Delta t_{log}))$\\
\STATE Clamp $a \in \left[a_{\min},a_{\max}\right]$, $b \in \left[b_{\min},b_{\max}\right]$ 
\FOR{$t$ \textnormal{in sequence}} 
\STATE $w \leftarrow \beta \cdot w + a \cdot u + b \cdot s$\\
\STATE $u \leftarrow \alpha \cdot (u - s) + (1 - \alpha) \cdot (I_t - w)$ \\
\STATE $s \leftarrow (u \geq \theta)$
\ENDFOR
\end{algorithmic}
\end{algorithm}
\subsection{C-SiLIF Model: SSM-Inspired Complex State}
State space models like S4 use complex-valued parameters and states, $x = x^{real} + i x^{img} \in \mathbb{C}$, which was recently shown to improve parameter efficiency and stability relative to real-valued parametrization \cite{ran2024provable}. 
A scalar complex transition $x_t = ax_{t-1} + bu_t$ with $a, b \in \mathbb{C}$, corresponds to the following 2x2 real-valued system: 
\begin{equation}
\begin{aligned}
\begin{bmatrix}x_t^{real}\\x_t^{img}\end{bmatrix}=\begin{bmatrix}a^{real}&-a^{img}\\a^{img}&a^{real}\end{bmatrix}\begin{bmatrix}x_{t-1}^{real}\\x_{t-1}^{img}\end{bmatrix} + \begin{bmatrix}b^{real}\\b^{img}\end{bmatrix}u_t
\end{aligned}
\label{eq:2by2sys}
\end{equation}
This has a similar format to the state transition in the AdLIF and SiLIF models, except that a complex system exhibits a structured, antisymmetric matrix. 
As demonstrated by \citet{ran2024provable}, such a complex parametrization allows for a more compact model with training-friendly, moderate-value parameters. 
We empirically corroborate the benefits of such parametrization on spiking neurons using event-based datasets (Sec.~\ref{sec:results}). 

Inspired by this complex constrained approach, we propose the complex-valued SSM-inspired LIF (C-SiLIF). C-SiLIF employs a complex state and follows the same dynamics as the 2x2 system described in Eq.~\ref{eq:2by2sys}. 
While S4D \cite{S4D} typically freezes the input projection $b = 1 + i$ and optimizes the output projection $\mathbf{\bar{C}}$ during training, we empirically found that freezing $\mathbf{\bar{C}} = 2$ and optimizing $b \in \mathbb{R}$ yields better performance for our model.
As in SiLIF, we incorporate a reset by introducing feedback from spiking events into the neuronal state. Since the output consists of twice the real part of the state variable, half the spike value is fed back into the state space to maintain consistent pre-synaptic dynamics.

Following the same reasoning as for the SiLIF model, we use heterogeneous parameters, logarithmic reparametrization, and timestep training for the transition parameters $\alpha^{real}$ and $\alpha^{img}$ to enhance the model's stability and capacity.
Additionally, for the C-SiLIF, we utilize S4D-Lin initialization derived from orthogonal polynomial projections for optimal state dynamics \citep{S4D_param}. \\
The final version of our complex-valued diagonal SSM-inspired LIF (C-SiLIF) is presented in Alg. \ref{alg:dsiLIF_alg}. 
While C-SiLIF closely resembles the Resonate and Fire (RF) model \citep{Izhikevich2001}, the novel parametrization, initialization, and discretization significantly enhance its stability \citep{zucchet2024}, dynamical richness (Fig.~\ref{fig:EVs}), and performance (Sec.~\ref{sec:results}) compared to prior RF models.
\begin{algorithm}[htbp!]
\footnotesize
\caption{C-SiLIF: SiLIF with complex state}\label{alg:dsiLIF_alg}
\begin{algorithmic}
    \STATE \textbf{Initialize trainable parameters:} 
    \STATE $\lambda^{real}_{log} \leftarrow  \log(0.5)$,  $\lambda^{img} \leftarrow \pi$, 
    \STATE $\Delta t_{log}  \sim \mathcal{U}(\log(\Delta t_{\text{min}}), \log(\Delta t_{\text{max}}))$, $b \sim \mathcal{U}(0, 1)$ 
    \STATE \textbf{Initialize states:} \\ $u \sim \mathcal{U}(0, 1)$, $s \sim \mathcal{U}(0, 1)$
    \STATE\vspace{-.7em}
    \STATE \textbf{Forward pass:} \\
    \STATE $\alpha \leftarrow \exp\left(\left(-\exp(\lambda^{real}_{log}) + i \cdot \lambda^{img}\right) \cdot \exp(\Delta t_{log})\right)$
    \FOR{$t$ \textnormal{in sequence}}
    \STATE $u \leftarrow \alpha \cdot (u - 0.5 \cdot s) + b \cdot I_t$
    \STATE $s \leftarrow (2 \mathrm{Re}(u) \geq \theta)$
    \ENDFOR
\end{algorithmic}
\end{algorithm}
\subsection{Neuronal Dynamics Regimes}
\label{sec:dynamics}
\begin{figure}[h]
    \centering
    \includegraphics[width=0.99\linewidth]{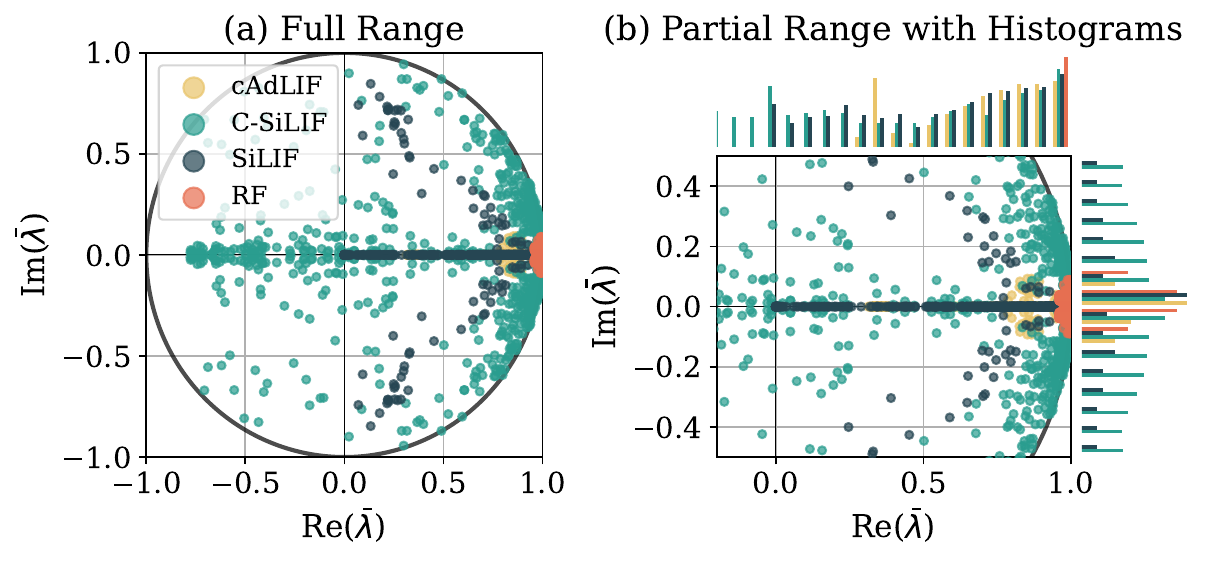}
    \caption{Scatter plot and histogram of state transition matrix eigenvalues for different pre-trained neuron models on the SSC dataset. (a) depicts the full unit circle, while (b) shows a zoomed view with a logarithmic range histogram of the eigenvalues. The range of covered eigenvalues, especially out of the real axis, corresponds to different accessible neuronal dynamics regimes.}
    \label{fig:EVs}
\end{figure}

We now examine the impact of the SSM-inspired features on the dynamics of our C-SiLIF and SiLIF models. We represent the distribution of eigenvalues of the transition matrix $\mathbf{\bar{A}}$ in the complex plane and compare these to the RF model and the cAdLIF \cite{co-learning-delays}. 
The cAdLIF simply constrains the adaptive parameter $a$ to positive values (Eq.~\ref{eq:adlif}), which helps stability and yields improved performance. 
The distributions are obtained after training models on the SSC dataset (see Sec.~\ref{sec:results}) and presented in Fig.~\ref{fig:EVs}. 

The cAdLIF model displays the most constrained eigenvalues, with mostly real values above $0.5$. 
In fact, cAdLIF and AdLIF use strict constraints on their parameters range, with decays $\alpha$ and $\beta$ slightly below 1 and the adaptation $a$ within $[-1,1]$, which mathematically results in predominantly real eigenvalues.  
The RF model, with its complex-constrained 2x2 system and fixed timestep $\Delta t$, yields a very concentrated eigenvalue distribution around 1 with limited imaginary components. 

On the other hand, our SiLIF and C-SiLIF models cover a considerably wider spectrum of neuronal dynamics:
The SiLIF covers most of the real-positive unit circle due to the absence of clamping on the transition parameters $\alpha$ and $\beta$ and the $\Delta t$ training, coinciding with significant performance improvements. 
Additionally, the complex system defined by the C-SiLIF allows transition parameters to be negative, leading to an eigenvalue distribution covering the entire unit circle. 
This allows our models to produce a wider, more flexible range of trained neuronal dynamics, including oscillatory ones which the cAdLIF or AdLIF typically cannot, as shown in the Appendix \ref{sec:regimes}.

We demonstrate in the Experiments section below that these widened neuronal dynamics arising from the $\Delta t$ training, the absence of parameter clamping, or the specific complex initialization significantly improve performance on audio classification tasks, establishing new SOTA results with our SSM-inspired models.
\section{Experiments}
\subsection{Experimental Setup}
\label{sec:setup}
\begin{table*}[!t]
\centering
\caption{Classification accuracy on SSC and GSC datasets.
Models are ranked by Top-1 accuracy (highest at the bottom) within non-delay and delay-based categories.
Our models are shown in bold. Non-delay and delay (Del.) models are separated by a horizontal line for each task. Models using inter-neuron recurrent connections are indicated as Rec.}
\begin{tabular}{llcccccc}
\toprule
\textbf{Dataset} & \textbf{Method} & \textbf{Resolution} & \textbf{\# Params} 
& \textbf{Rec.} & \textbf{Del.} & \textbf{Top1 Accuracy} \\
\midrule
\multirow{11}{*}{SSC}
& RadLIF \cite{Bittar2022} & 14\,ms & 3.9M & \checkmark &  & 77.40\% \\
& cAdLIF \cite{co-learning-delays} & 10\,ms & 0.35M &  &  & 77.50\% \\
& S5-RF \cite{huber2025} & 4\,ms & 1.7M &  &  & 78.8\% \\
& \textbf{SiLIF, 10\,ms bins (ours)} & \textbf{10\,ms} & \textbf{0.35M} &  &  & \textbf{80.11 $\pm$ 0.31\%} \\
& SE-adLIF \cite{baronig2025} & 4\,ms & 1.6M &  &  & 80.44 $\pm$ 0.26\% \\
& \textbf{C-SiLIF, 4\,ms bins (ours)} & \textbf{4\,ms} & \textbf{0.35M} &  &  & \textbf{81.59 $\pm$ 0.31\%} \\
& \textbf{SiLIF, 4\,ms bins (ours)} & \textbf{4\,ms} & \textbf{0.35M} &  &  & \textbf{82.03 $\pm$ 0.25\%} \\
\cline{2-7}
& d-cAdLIF \cite{co-learning-delays} & 10\,ms & 0.7M &  & \checkmark & 80.23 $\pm$ 0.07\% \\
& DCLS-Delays \cite{hammouamri2024} & 10\,ms & 1.2M &  & \checkmark & 80.29 $\pm$ 0.06\% \\
& \textbf{SiLIF + Delays (ours)} & \textbf{10\,ms} & \textbf{0.7M} &  & \checkmark & \textbf{80.62 $\pm$ 0.16\%} \\
& DelRec \cite{queant2025delrec} & 5.6\,ms & 0.37M & \checkmark & \checkmark & 82.58 $\pm$ 0.08\% \\
\midrule
\multirow{7}{*}{GSC}
& RadLIF \cite{Bittar2022} & 10\,ms & 0.83M & \checkmark &  & 94.51\% \\
& cAdLIF \cite{co-learning-delays} & 10\,ms & 0.3M &  &  & 94.67\% \\
& \textbf{SiLIF, 512 hidden units (ours)} & \textbf{10\,ms} & \textbf{0.3M} &  &  & \textbf{95.25 $\pm$ 0.12\%} \\
& \textbf{SiLIF, 1024 hidden units (ours)} & \textbf{10\,ms} & \textbf{1.1M} &  &  & \textbf{95.49 $\pm$ 0.09\%} \\
\cline{2-7}
& DCLS-Delays \cite{hammouamri2024} & 10\,ms & 1.2M &  & \checkmark & 95.29 $\pm$ 0.11\% \\
& d-cAdLIF \cite{co-learning-delays} & 10\,ms & 0.61M &  & \checkmark & 95.69 $\pm$ 0.03\% \\
& \textbf{SiLIF + Delays (ours)} & \textbf{10\,ms} & \textbf{0.61M} &  & \checkmark & \textbf{95.88 $\pm$ 0.12\%} \\
\bottomrule
\end{tabular}
\label{table:results_ssc_gsc}
\end{table*}
\begin{figure*}[h]
\centering
    \includegraphics[width=\textwidth]{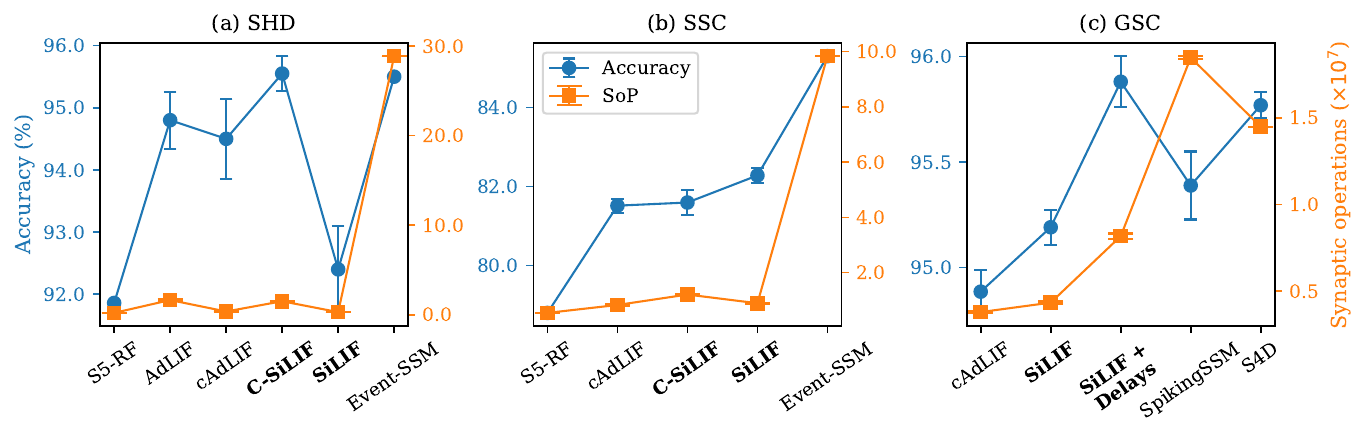}
\caption{Test accuracy and synaptic operations (SOP) with standard deviation for different models on the 3 audio datasets SHD, SSC, and GSC from left to right. Our models are shown in bold.}
\label{fig:sop}
\end{figure*}
We evaluate our models on keyword classification tasks, which are common benchmarks to evaluate spiking and recurrent neural networks. We use the Spiking Heidelberg Digits (SHD), the Spiking Speech Commands (SSC) \cite{cramer2020heidelberg}, and the Google Speech Commands v0.02 (GSC) dataset~\cite{warden2018speech}, in accordance with recent research on neuromorphic models \cite{Bittar2022, co-learning-delays, baronig2025, huber2025, hammouamri2024, schone2024, queant2025delrec}. 
GSC contains over 100k samples across 35 classes of spoken command words. 
SSC is an event-based converted version of GSC using a cochlea model \cite{cramer2020heidelberg}.
The SHD dataset consists of 10k audio samples, equally distributed across 20 classes of German and English spoken digits, and is converted to events similarly to SSC.

Different studies have used varying temporal binning and resolution for the spiking datasets, which has a significant impact on classification performance. 
Thus, we handle multiple cases to present fair comparisons: by default, we employ bins of $4\,ms$ for event-based datasets, and scale down to $10\,ms$ bins for comparison when required. 
Additionally, the spiking datasets are binned spatially from 700 to 140 input channels, matching recent studies. 
For GSC, the samples are fixed to $10\,ms$ time bins, obtained from a Mel filterbank with 40 Mel filters, matching the setup used in all similar studies.
Finally, we note that the SHD dataset lacks a separate test and validation set, which led previous studies to report the model's validation accuracy at the best epoch. 
We follow the same approach here for fair comparison. 

Our implementation builds on the SpArch\footnote{\url{https://github.com/idiap/sparch}, released under BSD-3-Clause License.} code by \citet{Bittar2022}. 
The corresponding network architecture has since been employed by most works in this field \citep{co-learning-delays, hammouamri2024, huber2025}. 
The network consists of two non-recurrent hidden layers with batch normalization and dropout. The layer width is fixed to 512 units unless otherwise specified. The output layer is fixed to a non-firing Leaky Integrate (LI) neuron whose outputs are passed through a softmax and summed over time. 
For our SiLIF and C-SiLIF models, the trainable neuron parameters and their initializations are listed in Algorithms \ref{alg:siLIF_alg} and \ref{alg:dsiLIF_alg}, respectively. 
The spiking threshold is fixed at 1. 
We use a cross-entropy loss with an Adam optimizer and a Plateau scheduler. 
We select hyperparameters for our models based on grid searches of the learning rate, dropout, and scheduler (see Appendix \ref{section:hyperparameters} for details). 
Details of other parameters, such as initialization ranges, can be found directly in our code.

In addition to the base SiLIF model introduced here, we implement a variant with learnable synaptic delays, termed "SiLIF + Delays", in which each synapse can delay incoming spikes by a certain number of timesteps before propagating them through the network. 
This variant is implemented separately following the DCLS-Delay framework and its proposed training pipeline \citep{hammouamri2024} (see Appendix \ref{sec:delays} for details).
Our reported results are averaged over 5 seeds for SHD, and 10 seeds for SSC and GSC.
The average $\mu$ and the standard deviation $\sigma$ are reported as $\mu \pm \sigma \%$.

Altogether this project required about 5000 GPU hours on a Nvidia RTX 4090 with 24 GB memory with a shared 377 GB AMD EPYC 9384X 32-Core CPU on an internal cluster, also considering initial exploratory runs. 
Each run takes an average of 2 hours for the SHD dataset, 15 hours for SSC, and 5 hours for GSC. 
This allowed for a detailed ablation study and reproduction of previous results with grid search analysis, targeting the most detailed and representative results possible. 

\subsection{Results on Audio Keyword Classification Tasks}
\label{sec:results}
\label{sec:sota}
We present results of our C-SiLIF, SiLIF, and SiLIF + Delays models on the SSC and GSC datasets in Table~\ref{table:results_ssc_gsc} and compare these to the previous state-of-the-art (SOTA) results obtained with reset-based spiking neuron models. For conciseness, the results for the smaller SHD dataset are presented in the Appendix \ref{sec:shd}.
Along with accuracy, we report the total number of parameters and indicate which models have recurrent connections (Rec.) and which ones have delays (Del.).
These two specific features typically improve performance but come at the cost of extra memory or computation, especially when implemented on neuromorphic hardware.

As presented in Table~\ref{table:results_ssc_gsc}, our models achieve significant improvements over previous SOTA models in all categories. 
The SiLIF model establishes new SOTA performances for non-delay models on both SSC and GSC by clear margins, while the version with synaptic delays establishes a new general SOTA for GSC with spiking neural networks. Our SiLIF + Delays model also achieves a new best performance for SSC at a $10\,ms$ resolution, while our models are only surpassed by the recent work from \citet{queant2025delrec}, which requires an increased resolution of $5.6\,ms$ along with costly recurrent connections.

\begin{table*}[h]
\centering
\caption{Test accuracy on the SHD dataset with ablation of SSM-inspired features of the C-SiLIF model. The configurations represent the best accuracy hyperparameters obtained from a 5-seed sweep for each model with 2 hidden layers of 512 neurons.}
\begin{tabular}{lccc}
    \toprule
    Model & Accuracy & Configuration \\
    \midrule
    \textbf{C-SiLIF} & 95.6 $\pm$ 0.3\% & lr=0.005, dropout=0.1\\
    C-SiLIF w/o half reset and input gating & 94.9 $\pm$ 0.3\% & lr=0.005, dropout=0.1 \\
    C-SiLIF w/o half reset, input gating, and S4D-Lin init. & 93.3 $\pm$ 0.5\% & lr=0.001, dropout=0.5 \\
    C-SiLIF w/o half reset, input gating, S4D-Lin init. and log. reparametrization & 90.2 $\pm$ 0.6\% & lr=0.001, dropout=0.1 \\
    C-SiLIF with $\Delta t$ constant (uniform distribution) & 95.1 $\pm$ 0.7\% & lr=0.01, dropout=0.5 \\
    C-SiLIF with $\Delta t$ constant (single value) & 89.3 $\pm$ 1.2\% & lr=0.005, dropout=0.1 \\
    \bottomrule
\end{tabular}
\label{table:ablation}
\end{table*}
\subsection{Computational Cost vs. Accuracy Analysis}
\label{sec:sparsity}
A key motivation for spiking neurons is their potential for reduced latency and energy consumption on dedicated event-based hardware \cite{Mead90_neurelec,Indiveri_etal11_neursili,Gao_etal18_deltpowe,Merolla_etal14_millspik, Pei_etal19_towaarti}, due to their spike-induced sparsity. 
Here, we assess model efficiency by reporting the number of synaptic operations (SOPs) per sample. 
The SOPs are computed here as the number of times a non-zero activity passes through a synaptic weight.

We report accuracy and SOPs from reproduced experiments on the AdLIF \citep{Bittar2022}, cAdLIF \cite{co-learning-delays}, and our C-SiLIF, SiLIF, and SiLIF + Delays models. 
All architectures consist of 2 hidden layers of 512 neurons and are trained without activity regularization loss or any other sparsity-enhancing method.
Each reported performance is obtained with 5 random seeds after a grid search.

In our efficiency comparisons with delay-based models, we assume that storing and loading delayed spikes is as costly as accumulating weights.
This leads to doubling the number of synaptic operations (SOPs).

Additionally, we compare these models to other SSM-related models. 
On the spiking datasets SHD and SSC, we also reported results from the S5-RF \citep{huber2025} and Event-SSM \citep{schone2024}.
The S5-RF is a SSM-inspired spiking neuron model employing the HIPPO initialization but no further reparametrization, which is a key feature of our work. 
\citet{huber2025} report SOPs per timestep; we thus scale them by the number of timesteps per sample to match our metric. 
The Event-SSM is a linear model made to process event streams, but without any spiking or reset mechanism, i.e., without any sparsity. 
Their SOP value computation is detailed in Appendix \ref{sec:ssm_ops}. 

\begin{figure}[h]
\begin{center}
    \includegraphics[width=0.99\linewidth]{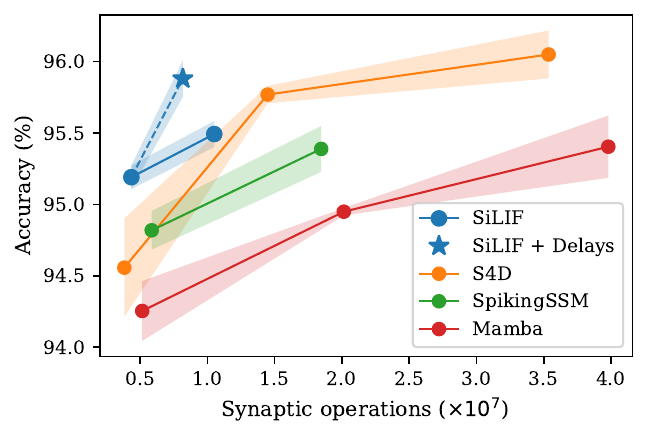}
\end{center}
\caption{Test accuracy per synaptic operations (SOPs) on the GSC dataset for our SiLIF model and SOTA SSM models at different scales. The standard deviation is represented as a shaded area.}
\label{fig:pareto_curve}
\end{figure}

On the GSC task, we report results for the standard S4D \citep{S4D}, the recent SpikingSSM model \citep{shen2025spikingssms}, a sparsified S4D model with appended LIF neurons, and Mamba \cite{gu2024mamba}, an enhanced version of S4D with selective gating.
While \citet{shen2025spikingssms} and \citet{ding2025keyword} already report results for these models on GSC, the respective use of raw-audio data at very high resolution and of bidirectional models results in orders of magnitude higher SOPs. 
We thus retrain SpikingSSM, S4D, and Mamba using our training pipeline to ensure a fair comparison (see Appendix \ref{sec:ssm_opti} for details). 
While we evaluate all three across multiple computational scales (Fig.~\ref{fig:pareto_curve}), only representative configurations of S4D and SpikingSSM are reported in Fig.~\ref{fig:sop}, as Mamba is consistently dominated in the accuracy-efficiency trade-off.

Fig.~\ref{fig:sop} summarizes the resulting accuracy and computational efficiency across the considered models.
Since the SHD and SSC experiments here are conducted at a $4\,ms$ resolution, we confirm that our C-SiLIF and SiLIF models, respectively, outperform all reset-based models reproduced. 
Our C-SiLIF is Pareto-optimal in accuracy-efficiency on the SHD task, while SiLIF and SiLIF + Delays are for the SSC and GSC tasks. 
Despite achieving top accuracy on the SSC task, the Event-SSM requires nearly an order of magnitude more operations and relies on dense real-valued activations. 
On the GSC dataset, despite its sparsity, SpikingSSM is outperformed by S4D at iso-compute. 
Both these linear recurrent models are outperformed by our SiLIF + Delays model at half the computational cost, while the standard SiLIF remains Pareto-optimal at a third of S4D's budget.

Further on, Fig.~\ref{fig:pareto_curve} shows the scaling curves of S4D, SpikingSSM, and Mamba next to our SiLIF model, scaled from 512 to 1024 hidden units and with added synaptic delays. We observe poor performance from Mamba, which is outperformed by all other models, potentially due to its optimization for large-scale models and long-sequence tasks. While SpikingSSM benefits from sparsity, the associated accuracy degradation makes it less competitive than S4D and SiLIF. Our SiLIF + Delays model shows great performance and is only marginally surpassed by the large S4D configuration, while requiring four times fewer operations. \\
Altogether, these results firmly establish the competitive edge of our reset-based spiking models over SOTA SNN and SSM baselines for audio tasks on neuromorphic hardware.
\subsection{Ablation Study}
\label{sec:ablation}
We conduct an ablation study of the C-SiLIF model, considering that it includes all SiLIF features plus the complex-value constraint and S4D initialization. 
Each ablation configuration is optimized with a hyperparameter grid search to reach optimized performance for this configuration. 
We report the best obtained performance along with the corresponding hyperparameter configuration in Table~\ref{table:ablation}.

The ablation mostly highlights the importance of the logarithmic reparametrization and the timestep $\Delta t$, which are common features to our two models. 
Removing reparametrization from lines 3-4 of Table~\ref{table:ablation} results in a performance drop of more than $3\%$. 
Additionally, we show for the first time the importance of a heterogeneous, or even trained $\Delta t$ for reset-based spiking neurons, especially as the use of a single homogeneous $\Delta t$ on the last line leads to more than $6\%$ performance drop. 
As seen in Fig.~\ref{fig:EVs}, this directly impacts the range of transition parameters and thus dynamic regimes covered by the model. This underscores the correlation between dynamic regimes and task performance. 
We also study the impact of these different features in an incremental fashion in Appendix \ref{sec:incr_abl}.

\section{Conclusion}
This work presents a theoretical bridge between state space models (SSMs) and spiking neurons and proposes two novel spiking neuron models integrating SSM-inspired dynamics and parametrization. 
In particular, we show the importance of heterogeneous timestep parameters $\Delta t$ along with the use of logarithmic reparametrization, which leads to a wider range of dynamical regimes and, eventually, to improved performance. 
Our models achieve new state-of-the-art (SOTA) performances across the board on speech recognition tasks, both for delay and non-delay models, with significant margins and reduced scale.
The SiLIF benefits also extend to non-reset models, where our model is Pareto-optimal in performance-efficiency with respect to other SSM-SNN hybrids and even modern SSM models, opening a new avenue for neuromorphic models and systems at the edge.
Future work could investigate the use of SSM state expansion in spiking neurons or apply our method to other network architectures or sensory processing tasks.

\section*{Acknowledgments}
This work was sponsored by the Federal Ministry of Education, Germany (project NEUROTEC-II grant no. 16ME0398K and 16ME0399),  by Neurosys as part of the initiative "Cluster4Future" is funded by the Federal Ministery of Education and Research BMBF (03ZU1106CB) and by the Federal Ministry of Education and Research (BMBF) under grant no. 01IS22094E WEST-AI. The authors also extend their gratitude to Professor Elisabetta Chicca for her general feedback on analog hardware implications and Sanja Karilanova for our insightful discussions on the link between state space models and spiking neurons.

\section*{Impact Statement}
This paper presents work whose goal is to advance the field of Machine Learning. There are many potential societal consequences of our work, none of which we feel must be specifically highlighted here.

\pagebreak

\bibliography{main_refs,biblio_unique_alt}

@InProceedings{Gao_etal18_deltpowe,
author		= {Gao, Chang and Neil, Daniel and Ceolini, Enea and Liu,
		  ShihChii and Delbruck, Tobi},
title		= {DeltaRNN: A Powerefficient Recurrent Neural Network
		  Accelerator},
year		= {2018},
isbn		= {9781450356145},
publisher	= {Association for Computing Machinery},
address		= {New York, NY, USA},
url		= {https://doi.org/10.1145/3174243.3174261},
doi		= {10.1145/3174243.3174261},
booktitle	= {Proceedings of the 2018 ACM/SIGDA International Symposium
		  on FieldProgrammable Gate Arrays},
pages		= {21–30},
numpages	= {10},
location	= {Monterey, CALIFORNIA, USA},
series		= {FPGA '18}
}

@Article{Indiveri_etal11_neursili,
author		= {G. Indiveri and B. LinaresBarranco and T.J. Hamilton and
		  A. van Schaik and R. EtienneCummings and T. Delbruck and
		  S.C. Liu and P. Dudek and P. H{\"a}fliger and S. Renaud and
		  J. Schemmel and G. Cauwenberghs and J. Arthur and K. Hynna
		  and F. Folowosele and S. Saighi and T. SerranoGotarredona
		  and J. Wijekoon and Y. Wang and K. Boahen},
doi		= {10.3389/fnins.2011.00073},
issn		= {1662453X},
journal		= {Frontiers in Neuroscience},
link		= {http://www.frontiersin.org/Neuromorphic_Engineering/10.3389/fnins.2011.00073/abstract},
pages		= {123},
title		= {Neuromorphic silicon neuron circuits},
volume		= {5},
year		= {2011}
}

@Article{Mead90_neurelec,
author		= {C. Mead},
journal		= {Proceedings of the {IEEE}},
number		= {10},
pages		= {162936},
title		= {Neuromorphic Electronic Systems},
volume		= {78},
year		= {1990}
}

@Article{Merolla_etal14_millspik,
author		= {Merolla, Paul A and Arthur, John V and AlvarezIcaza,
		  Rodrigo and Cassidy, Andrew S and Sawada, Jun and Akopyan,
		  Filipp and Jackson, Bryan L and Imam, Nabil and Guo, Chen
		  and Nakamura, Yutaka and others},
journal		= {Science},
number		= {6197},
pages		= {668673},
publisher	= {American Association for the Advancement of Science},
title		= {A million spikingneuron integrated circuit with a scalable
		  communication network and interface},
volume		= {345},
year		= {2014}
}

@Article{Pei_etal19_towaarti,
title		= {Towards artificial general intelligence with hybrid
		  Tianjic chip architecture},
author		= {Pei, Jing and Deng, Lei and Song, Sen and Zhao, Mingguo
		  and Zhang, Youhui and Wu, Shuang and Wang, Guanrui and Zou,
		  Zhe and Wu, Zhenzhi and He, Wei and others},
journal		= {Nature},
volume		= {572},
number		= {7767},
pages		= {106111},
year		= {2019},
publisher	= {Nature Publishing Group UK London}
}

@ARTICLE{Neftci19,
  author={Neftci, Emre O. and Mostafa, Hesham and Zenke, Friedemann},
  journal={IEEE Signal Processing Magazine}, 
  title={Surrogate Gradient Learning in Spiking Neural Networks: Bringing the Power of Gradient-Based Optimization to Spiking Neural Networks}, 
  year={2019},
  volume={36},
  number={6},
  pages={51-63},
  doi={10.1109/MSP.2019.2931595}}

@article{Lee16,
   author = {Jun Haeng Lee and Tobi Delbruck and Michael Pfeiffer},
   doi = {10.3389/fnins.2016.00508},
   issn = {1662453X},
   issue = {NOV},
   journal = {Frontiers in Neuroscience},
   title = {Training deep spiking neural networks using backpropagation},
   volume = {10},
   year = {2016},
}

@inproceedings{Bu2022,
   author = {Tong Bu and Wei Fang and Jianhao Ding and Peng Lin Dai and Zhaofei Yu and Tiejun Huang},
   booktitle = {ICLR 2022 - 10th International Conference on Learning Representations},
   title = {OPTIMAL ANN-SNN CONVERSION FOR HIGH-ACCURACY AND ULTRA-LOW-LATENCY SPIKING NEURAL NETWORKS},
   year = {2022},
}

@inproceedings{hippo,
  title={HiPPO: Recurrent Memory with Optimal Polynomial Projections},
  author={Gu, Albert and Dao, Tri and Ermon, Stefano and Rudra, Atri and R{\'e}, Christopher},
  booktitle={Advances in Neural Information Processing Systems},
  volume={33},
  year={2020},
  url={https://arxiv.org/abs/2008.07669}
}

@inproceedings{S4D,
   author = {Ankit Gupta and Albert Gu and Jonathan Berant},
   issn = {10495258},
   booktitle = {Advances in Neural Information Processing Systems},
   title = {Diagonal State Spaces are as Effective as Structured State Spaces},
   volume = {35},
   year = {2022},
}

@inproceedings{S4D_param,
   author = {Albert Gu and Ankit Gupta and Karan Goel and Christopher Ré},
   issn = {10495258},
   booktitle = {Advances in Neural Information Processing Systems},
   title = {On the Parameterization and Initialization of Diagonal State Space Models},
   volume = {35},
   year = {2022},
}

@article{Brette2005,
   author = {Romain Brette and Wulfram Gerstner},
   doi = {10.1152/jn.00686.2005},
   issn = {00223077},
   issue = {5},
   journal = {Journal of Neurophysiology},
   title = {Adaptive exponential integrate-and-fire model as an effective description of neuronal activity},
   volume = {94},
   year = {2005},
}

@article{Salaj2021,
   author = {Darjan Salaj and Anand Subramoney and Ceca Kraisnikovic and Guillaume Bellec and Robert Legenstein and Wolfgang Maass},
   doi = {10.7554/eLife.65459},
   issn = {2050084X},
   journal = {eLife},
   title = {Spike frequency adaptation supports network computations on temporally dispersed information},
   volume = {10},
   year = {2021},
}

@article{Yin2021,
   author = {Bojian Yin and Federico Corradi and Sander M. Bohté},
   doi = {10.1038/s42256-021-00397-w },
   issn = {25225839},
   issue = {10},
   journal = {Nature Machine Intelligence},
   title = {Accurate and efficient time-domain classification with adaptive spiking recurrent neural networks},
   volume = {3},
   year = {2021},
}

@article{co-learning-delays,
  title={Co-learning synaptic delays, weights and adaptation in spiking neural networks},
  author={Deckers, Lucas and Van Damme, Laurens and Tsang, Ing Jyh and Van Leekwijck, Werner and Latré, Steven},
  journal={Frontiers in Neuroscience},
  volume={18},
  year={2024},
  doi={10.3389/fnins.2024.1360300}
}

@inproceedings{hammouamri2024,
  title={Learning Delays in Spiking Neural Networks using Dilated Convolutions with Learnable Spacings},
  author={Hammouamri, Ilyass and Khalfaoui-Hassani, Ismail and Masquelier, Timoth{\'e}e},
  booktitle={International Conference on Learning Representations},
  year={2024},
  url={https://openreview.net/forum?id=4r2ybzJnmN}
}

@article{Bittar2022,
   author = {Alexandre Bittar and Philip N. Garner},
   doi = {10.3389/fnins.2022.865897 },
   issn = {1662453X},
   journal = {Frontiers in Neuroscience},
   title = {A surrogate gradient spiking baseline for speech command recognition},
   volume = {16},
   year = {2022},
}

@article{baronig2025,
  title={Advancing spatio-temporal processing in spiking neural networks through adaptation},
  author={Baronig, Maximilian and Ferrand, Romain and Sabathiel, Silvester and Legenstein, Robert},
  journal={Nature Communications},
  year={2025},
  url={https://www.nature.com/articles/s41467-025-60878-z}
}

@article{Izhikevich2001,
   author = {Eugene M. Izhikevich},
   doi = {10.1016/S0893-6080(01)00078-8 },
   issn = {08936080},
   issue = {6-7},
   journal = {Neural Networks},
   title = {Resonate-and-fire neurons},
   volume = {14},
   year = {2001},
}

@article{Frady2022,
   author = {E. Paxon Frady and Sophia Sanborn and Sumit Bam Shrestha and Daniel Ben Dayan Rubin and Garrick Orchard and Friedrich T. Sommer and Mike Davies},
   doi = {10.1007/s11265-022-01772-5 },
   issn = {19398115},
   issue = {10},
   journal = {Journal of Signal Processing Systems},
   title = {Efficient Neuromorphic Signal Processing with Resonator Neurons},
   volume = {94},
   year = {2022},
}

@misc{alkhamissi2021,
      title={Deep Spiking Neural Networks with Resonate-and-Fire Neurons}, 
      author={Badr AlKhamissi and Muhammad ElNokrashy and David Bernal-Casas},
      year={2021},
      eprint={2109.08234},
      archivePrefix={arXiv},
      primaryClass={cs.NE},
      url={https://arxiv.org/abs/2109.08234}, 
}

@misc{higuchi2024,
      title={Balanced Resonate-and-Fire Neurons}, 
      author={Saya Higuchi and Sebastian Kairat and Sander M. Bohte and Sebastian Otte},
      year={2024},
      eprint={2402.14603},
      archivePrefix={arXiv},
      primaryClass={cs.NE},
      url={https://arxiv.org/abs/2402.14603}, 
}

@misc{S5,
      title={Simplified State Space Layers for Sequence Modeling}, 
      author={Jimmy T. H. Smith and Andrew Warrington and Scott W. Linderman},
      year={2023},
      booktitle={International Conference on Neuromorphic Systems},
      url={https://openreview.net/pdf?id=Ai8Hw3AXqks}, 
}

@inproceedings{S4,
   author = {Albert Gu and Karan Goel and Christopher Ré},
   booktitle = {ICLR 2022 - 10th International Conference on Learning Representations},
   title = {EFFICIENTLY MODELING LONG SEQUENCES WITH STRUCTURED STATE SPACES},
   year = {2022},
}

@misc{neurobench,
      title={NeuroBench: A Framework for Benchmarking Neuromorphic Computing Algorithms and Systems}, 
      author={Jason Yik and Korneel Van den Berghe and Douwe den Blanken and Younes Bouhadjar and Maxime Fabre and Paul Hueber and Weijie Ke and Mina A Khoei and Denis Kleyko and Noah Pacik-Nelson and Alessandro Pierro and Philipp Stratmann and Pao-Sheng Vincent Sun and Guangzhi Tang and Shenqi Wang and Biyan Zhou and Soikat Hasan Ahmed and George Vathakkattil Joseph and Benedetto Leto and Aurora Micheli and Anurag Kumar Mishra and Gregor Lenz and Tao Sun and Zergham Ahmed and Mahmoud Akl and Brian Anderson and Andreas G. Andreou and Chiara Bartolozzi and Arindam Basu and Petrut Bogdan and Sander Bohte and Sonia Buckley and Gert Cauwenberghs and Elisabetta Chicca and Federico Corradi and Guido de Croon and Andreea Danielescu and Anurag Daram and Mike Davies and Yigit Demirag and Jason Eshraghian and Tobias Fischer and Jeremy Forest and Vittorio Fra and Steve Furber and P. Michael Furlong and William Gilpin and Aditya Gilra and Hector A. Gonzalez and Giacomo Indiveri and Siddharth Joshi and Vedant Karia and Lyes Khacef and James C. Knight and Laura Kriener and Rajkumar Kubendran and Dhireesha Kudithipudi and Shih-Chii Liu and Yao-Hong Liu and Haoyuan Ma and Rajit Manohar and Josep Maria Margarit-Taulé and Christian Mayr and Konstantinos Michmizos and Dylan R. Muir and Emre Neftci and Thomas Nowotny and Fabrizio Ottati and Ayca Ozcelikkale and Priyadarshini Panda and Jongkil Park and Melika Payvand and Christian Pehle and Mihai A. Petrovici and Christoph Posch and Alpha Renner and Yulia Sandamirskaya and Clemens JS Schaefer and André van Schaik and Johannes Schemmel and Samuel Schmidgall and Catherine Schuman and Jae-sun Seo and Sadique Sheik and Sumit Bam Shrestha and Manolis Sifalakis and Amos Sironi and Kenneth Stewart and Matthew Stewart and Terrence C. Stewart and Jonathan Timcheck and Nergis Tömen and Gianvito Urgese and Marian Verhelst and Craig M. Vineyard and Bernhard Vogginger and Amirreza Yousefzadeh and Fatima Tuz Zohora and Charlotte Frenkel and Vijay Janapa Reddi},
      year={2025},
      eprint={2304.04640},
      archivePrefix={arXiv},
      primaryClass={cs.AI},
      url={https://arxiv.org/abs/2304.04640}, 
}

@INPROCEEDINGS{Lyes,
  author={Caccavella, Caterina and Paredes-Vallés, Federico and Cannici, Marco and Khacef, Lyes},
  booktitle={2024 International Joint Conference on Neural Networks (IJCNN)}, 
  title={Low-power event-based face detection with asynchronous neuromorphic hardware}, 
  year={2024},
  volume={},
  number={},
  pages={1-10},
  doi={10.1109/IJCNN60899.2024.10650843}}

@inproceedings{statespacemodelsevent,
  title={State Space Models for Event Cameras},
  author={Zubi{\'c}, Nikola and Gehrig, Mathias and Scaramuzza, Davide},
  booktitle={IEEE/CVF Conference on Computer Vision and Pattern Recognition},
  year={2024},
  url={https://openaccess.thecvf.com/content/CVPR2024/papers/Zubic_State_Space_Models_for_Event_Cameras_CVPR_2024_paper.pdf}
}

@inproceedings{schone2024,
  title={Scalable Event-by-event Processing of Neuromorphic Sensory Signals with Deep State-Space Models},
  author={Sch{\"o}ne, Mark and Sushma, Neeraj Mohan and Zhuge, Jingyue and Mayr, Christian and Subramoney, Anand and Kappel, David},
  booktitle={International Conference on Neuromorphic Systems},
  year={2024},
  url={https://ieeexplore.ieee.org/document/10766552} 
}

@inproceedings{s6,
  title={P-SpikeSSM: Harnessing Probabilistic Spiking State Space Models for Long-Range Dependency Tasks},
  author={Bal, Malyaban and Sengupta, Abhronil},
  booktitle={International Conference on Learning Representations},
  year={2025},
  url={https://openreview.net/pdf?id=35gF1nqsgw}
}

@misc{huang2024prfparallelresonateneuron,
      title={PRF: Parallel Resonate and Fire Neuron for Long Sequence Learning in Spiking Neural Networks}, 
      author={Yulong Huang and Zunchang Liu and Changchun Feng and Xiaopeng Lin and Hongwei Ren and Haotian Fu and Yue Zhou and Hong Xing and Bojun Cheng},
      year={2024},
      eprint={2410.03530},
      archivePrefix={arXiv},
      primaryClass={cs.NE},
      url={https://arxiv.org/abs/2410.03530}, 
}

@inproceedings{zucchet2024,
  title={Recurrent neural networks: vanishing and exploding gradients are not the end of the story},
  author={Zucchet, Nicolas and Orvieto, Antonio},
  booktitle={Advances in Neural Information Processing Systems},
  year={2024},
  url={https://proceedings.neurips.cc/paper_files/paper/2024/file/fbb07254ef01868967dc891ea3fa6c13-Paper-Conference.pdf}
}

@article{queant2025delrec,
  title={Delrec: learning delays in recurrent spiking neural networks},
  author={Queant, Alexandre and Ran{\c{c}}on, Ulysse and Cottereau, Benoit R and Masquelier, Timoth{\'e}e},
  journal={arXiv preprint arXiv:2509.24852},
  year={2025}
}

@inproceedings{
    Hassani23_dcls,
    title={Dilated convolution with learnable spacings},
    author={Ismail Khalfaoui Hassani and Thomas Pellegrini and Timoth{\'e}e Masquelier},
    booktitle={The Eleventh International Conference on Learning Representations },
    year={2023},
    url={https://openreview.net/forum?id=Q3-1vRh3HOA}
}

@inproceedings{shen2025spikingssms,
  title={Spikingssms: Learning long sequences with sparse and parallel spiking state space models},
  author={Shen, Shuaijie and Wang, Chao and Huang, Renzhuo and Zhong, Yan and Guo, Qinghai and Lu, Zhichao and Zhang, Jianguo and Leng, Luziwei},
  booktitle={Proceedings of the AAAI Conference on Artificial Intelligence},
  volume={39},
  number={19},
  pages={20380--20388},
  year={2025}
}

@inproceedings{gu2024mamba,
  title={Mamba: Linear-time sequence modeling with selective state spaces},
  author={Gu, Albert and Dao, Tri},
  booktitle={First conference on language modeling},
  year={2024}
}

@InProceedings{huber2025,
author="Huber, Thomas E.
and Lecomte, Jules
and Polovnikov, Borislav
and von Arnim, Axel",
editor="Del Bue, Alessio
and Canton, Cristian
and Pont-Tuset, Jordi
and Tommasi, Tatiana",
title="Scaling Up Resonate-and-Fire Networks for Fast Deep Learning",
booktitle="Computer Vision -- ECCV 2024 Workshops",
year="2025",
publisher="Springer Nature Switzerland",
address="Cham",
pages="241--258",
isbn="978-3-031-92460-6"
}

@article{karilanova2025zero,
  title={Zero-shot temporal resolution domain adaptation for spiking neural networks},
  author={Karilanova, Sanja and Fabre, Maxime and Neftci, Emre and {\"O}z{\c{c}}elikkale, Ay{\c{c}}a},
  journal={Neural Networks},
  pages={108483},
  year={2025},
  publisher={Elsevier},
  url={https://www.sciencedirect.com/science/article/pii/S0893608025013644}
}

@inproceedings{orvieto2023resurrecting,
  title={Resurrecting recurrent neural networks for long sequences},
  author={Orvieto, Antonio and Smith, Samuel L and Gu, Albert and Fernando, Anushan and Gulcehre, Caglar and Pascanu, Razvan and De, Soham},
  booktitle={International Conference on Machine Learning},
  pages={26670--26698},
  year={2023},
  organization={PMLR},
  url={https://proceedings.mlr.press/v202/orvieto23a/orvieto23a.pdf}
}

@inproceedings{tay2020long,
  title={Long range arena: A benchmark for efficient transformers},
  author={Tay, Yi and Dehghani, Mostafa and Abnar, Samira and Shen, Yikang and Bahri, Dara and Pham, Philip and Rao, Jinfeng and Yang, Liu and Ruder, Sebastian and Metzler, Donald},
  booktitle={International Conference on Learning Representations},
  year={2021}
}

@article{kaiser2020synaptic,
  title={Synaptic plasticity dynamics for deep continuous local learning (DECOLLE)},
  author={Kaiser, Jacques and Mostafa, Hesham and Neftci, Emre},
  journal={Frontiers in Neuroscience},
  volume={14},
  pages={424},
  year={2020},
  publisher={Frontiers Media SA}
}

@article{zenke2018superspike,
  title={Superspike: Supervised learning in multilayer spiking neural networks},
  author={Zenke, Friedemann and Ganguli, Surya},
  journal={Neural computation},
  volume={30},
  number={6},
  pages={1514--1541},
  year={2018},
  publisher={MIT Press One Rogers Street, Cambridge, MA 02142-1209, USA journals-info~…}
}

@article{herranz2022stabilizing,
  title={Stabilizing spiking neuron training},
  author={Herranz-Celotti, Luca and Rouat, Jean},
  journal={arXiv preprint arXiv:2202.00282},
  year={2022}
}

@book{gerstner2014neuronal,
  title={Neuronal dynamics: From single neurons to networks and models of cognition},
  author={Gerstner, Wulfram and Kistler, Werner M and Naud, Richard and Paninski, Liam},
  year={2014},
  publisher={Cambridge University Press}
}

@article{ran2024provable,
  title={Provable benefits of complex parameterizations for structured state space models},
  author={Ran-Milo, Yuval and Lumbroso, Eden and Cohen-Karlik, Edo and Giryes, Raja and Globerson, Amir and Cohen, Nadav},
  journal={Advances in Neural Information Processing Systems},
  volume={37},
  pages={115906--115939},
  year={2024}
}

@article{cramer2020heidelberg,
  title={The heidelberg spiking data sets for the systematic evaluation of spiking neural networks},
  author={Cramer, Benjamin and Stradmann, Yannik and Schemmel, Johannes and Zenke, Friedemann},
  journal={IEEE Transactions on Neural Networks and Learning Systems},
  volume={33},
  number={7},
  pages={2744--2757},
  year={2020},
  publisher={IEEE}
}

@article{warden2018speech,
  title={Speech commands: A dataset for limited-vocabulary speech recognition},
  author={Warden, Pete},
  journal={arXiv preprint arXiv:1804.03209},
  year={2018}
}

@inproceedings{dao2024transformers,
  title={Transformers are SSMs: Generalized Models and Efficient Algorithms Through Structured State Space Duality},
  author={Dao, Tri and Gu, Albert},
  booktitle={International Conference on Machine Learning},
  pages={10041--10071},
  year={2024},
  organization={PMLR}
}

@article{beck2024xlstm,
  title={xlstm: Extended long short-term memory},
  author={Beck, Maximilian and P{\"o}ppel, Korbinian and Spanring, Markus and Auer, Andreas and Prudnikova, Oleksandra and Kopp, Michael and Klambauer, G{\"u}nter and Brandstetter, Johannes and Hochreiter, Sepp},
  journal={Advances in Neural Information Processing Systems},
  volume={37},
  pages={107547--107603},
  year={2024}
}

@article{ding2025keyword,
  title={Keyword Mamba: Spoken keyword spotting with state space models},
  author={Ding, Hanyu and Dong, Wenlong and Mao, Qirong},
  journal={Computer Speech \& Language},
  pages={101909},
  year={2025},
  publisher={Elsevier}
}
\bibliographystyle{icml2026}

\newpage
\appendix
\onecolumn
\section{Appendix}

\subsection{Link between Eigenvalues and Neuronal Dynamics Regimes}
\label{sec:regimes}
We give additional insights into the importance of the transition matrix eigenvalues of spiking neurons and their link to the dynamical regimes of the system.

The behavior of neurons with second order dynamics, including AdLIF, cAdLIF, C-SiLIF and SiLIF, can be classified as resonator, integrator, or mixed, depending on the ratio of the membrane and adaptation time constants (\(\tau_m / \tau_w = \beta / \alpha\))  and the coupling constant (\(a\)) (see Equation \ref{eq:adlif}). These parameters determine whether the system undergoes a Hopf or saddle-node bifurcation, which is determinant for a periodic system's stability. These parameters are thus typically clamped within specific ranges to control neuronal behavior \cite{Bittar2022}. 

\begin{figure}[h]
    \centering
    
    \subfloat[cAdLIF]{%
        \includegraphics[width=0.45\linewidth]{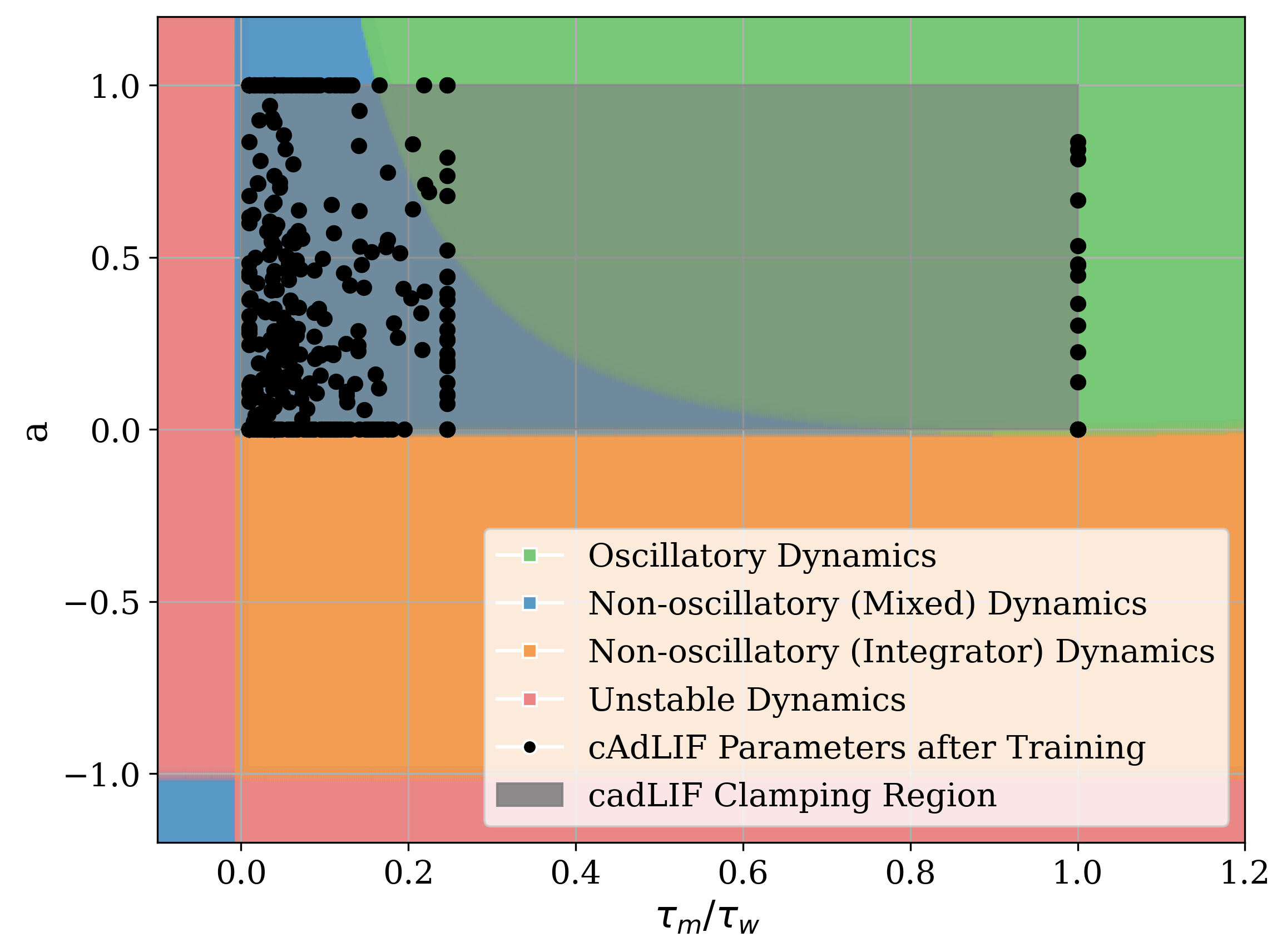}
        \label{fig:regions}
    }
    \subfloat[SiLIF]{%
        \includegraphics[width=0.45\linewidth]{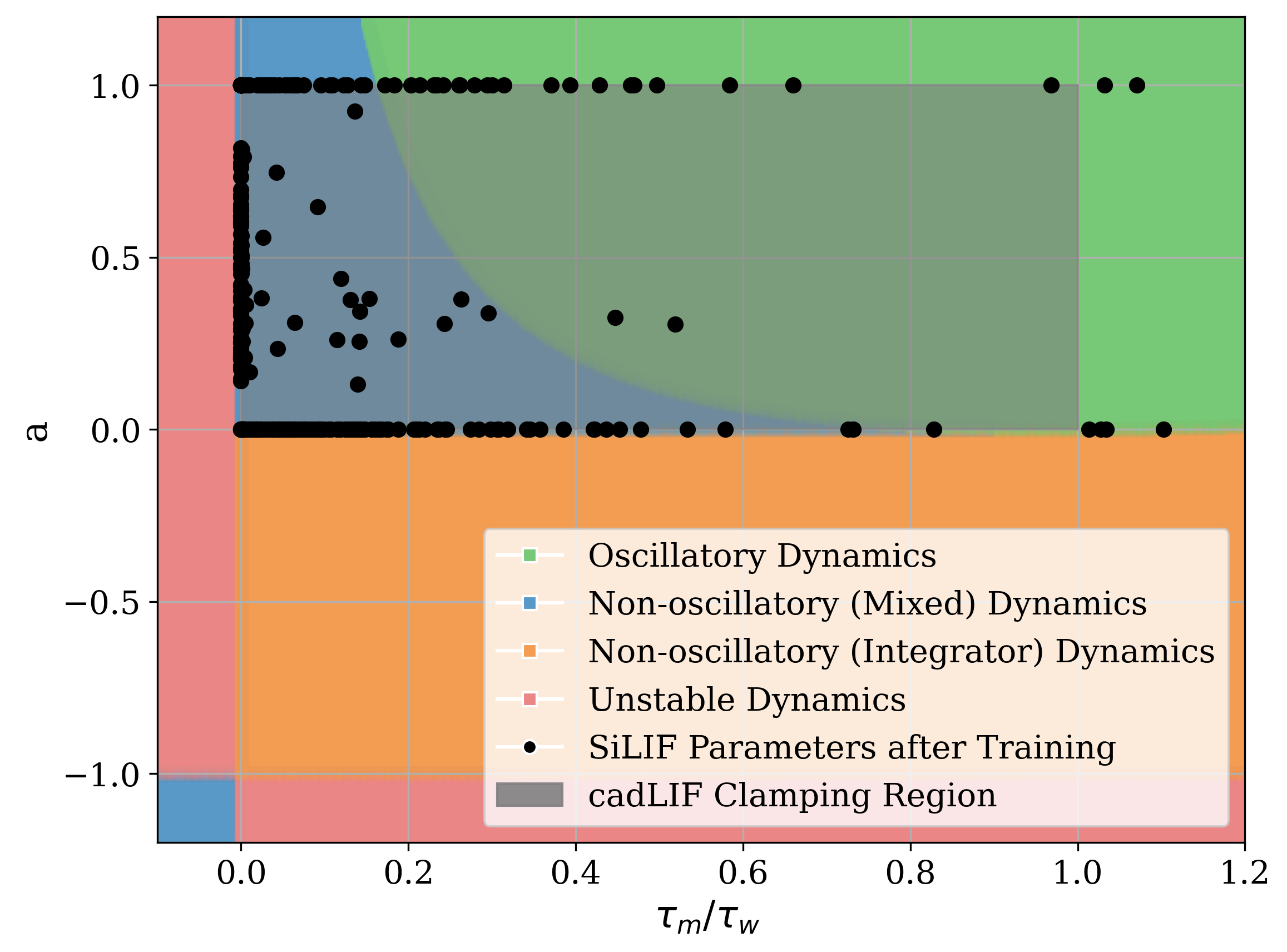}
    }
    \caption{Neuronal dynamics of the (a) cAdLIF and (b) SiLIF models pre-trained on the SSC task. Each dot corresponds to the obtained training parameters for one neuron of the model. The position of each dot directly correlates to the corresponding neuron's regime.}
    \label{fig:regimes}
\end{figure}

As seen in Figure \ref{fig:regimes}, cAdLIF and SiLIF models exhibit both integrator and resonator regimes. The resonator regime corresponds to complex eigenvalues with a non-zero imaginary part, while the integrator regime is associated with purely positive real eigenvalues, as illustrated in Figure \ref{fig:EVs}. Since $\alpha$ and $\beta$ are not explicitly clamped for the SiLIF model, oscillatory dynamics are more thoroughly explored by the neuron, as there is no artificial constraint on the $\beta/\alpha$ ratio. This leads to more parameter exploration flexibility during training and richer internal dynamics. Nevertheless, due to the exponential reparametrization, the SiLIF model always remains in the stable regime.

\subsection{Experimental Details of the SiLIF Model with Synaptic Delays}
\label{sec:delays}

\subsubsection{Synaptic Delays Model Description}
We adopt the convolution-based formulation of synaptic delays proposed in \cite{hammouamri2024}. Each individual synapse $(j \rightarrow i)$ is assigned its own learnable delay kernel, allowing different transmission latencies for each pre--post connection. The synaptic current received by neuron $i$ is written as
\begin{equation}
    I_i(t)=\sum_{j=1}^{N} w_{ij}\, S_j(t-d_{ij}) \;=\; \sum_{j=1}^{N} k_{ij} \ast S_j ,
\end{equation}
where $S_j \in \{0, 1\}$ is the spike or non-spike from neuron $j$, $w_{i,j}$ the synaptic weight, $d_{ij}$ the synaptic delay, $k_{ij}$ the corresponding kernel, and $\ast$ denotes discrete-time convolution.
Delays are learned using the one-dimensional DCLS parametrization of \cite{Hassani23_dcls}. In this formulation, each kernel $k_{ij}$ is modeled as a Gaussian with standard deviation $\sigma$, centered at $T_\text{d} - d_{ij} - 1$, where $T_{\text d}$ is the maximum allowable delay. The kernel length is set to $T_{\text d} + 1$. Consequently, each synapse is parameterized by a weight $w_{ij}$, a delay $d_{ij}$, and a standard deviation $\sigma$.
During training, all standard deviations $\sigma$ are initialized to $T_{\text d}/2$ and progressively annealed to $0.5$ over the first quarter of the training epochs. At test time, the learned delays are discretized by rounding to the nearest integer, following \cite{hammouamri2024}.

\subsubsection{Computational Cost of Synaptic Delays}
As mentioned in the main text, we note that delay-based models incur additional memory and computational costs when implemented on neuromorphic hardware. Indeed, synaptic delays require buffering incoming spikes for each synapse and delivering them later when the delay has passed. This mostly requires additional dynamic memory, but also some compute to store and load delayed spikes into buffers. 
Here, we consider the worst case scenario for delay-based models, where the storing and loading of delayed spikes is as costly as the accumulation of weights. This leads to doubling the number of synaptic operations (SOPs). Therefore, in Figures \ref{fig:sop} and \ref{fig:pareto_curve}, the reported SOPs for the SiLIF + Delays model are twice the number of effective SOPs for the synaptic weight accumulations.

\subsubsection{Experimental Setup for the SiLIF + Delays}
\par
We perform the experiments of the SiLIF model with synaptic delays using the code base by \citet{hammouamri2024} provided in \url{https://github.com/Thvnvtos/SNN-delays}.
The network architecture consists of 2 layers, each composed of a DCLS module, batch normalization, and a SiLIF layer with 512 neurons. The membrane potential of the leaky integrator is used for the readout mechanism. Key architectural hyperparameters are reported in \cref{tab:dcls_learning_parameters}.
The preprocessing of the SSC and GSC datasets closely follows the procedure described in \cref{sec:setup}. In particular, for GSC, we use 40 Mel-frequency channels, whereas the original work by \citet{hammouamri2024} setup uses 140 Mel-frequency channels.
After a hyperparameter search, detailed in \cref{tab:dcls_learning_parameters}, we find that the configurations yielding the best performance for synaptic delays closely match those reported in \cite{hammouamri2024}. In particular, we employ the Adam optimizer with separate parameter groups, using base learning rates of $10^{-3}$ for the synaptic weights and $10^{-1}$ for the delay parameters. A one-cycle learning rate scheduler is applied to the synaptic weights, while a cosine annealing schedule is used for the delay learning rates.

\begin{table*}[h]
    \centering
    \caption{Learning hyperparameter search spaces for the SiLIF + Delays model. Bold numbers indicate the selected hyperparameters. For the one-cycle scheduler, the maximum learning rate is set to $5\times$ the base learning rate, and the total number of steps in a cycle is set to the number of training epochs; for cosine annealing, the maximum number of iterations is also set to the number of training epochs.}

    \begin{tabular}{l c}
    \toprule
    \textbf{Hyperparameter} & \textbf{GSC/SSC} \\
    \midrule
    \texttt{$T_\text{d}$} & $\{5,\mathbf{11},15,31\}$ \\
    \texttt{$lr_\text{w}$ (weights)} & $\{10^{-2},\,\mathbf{10^{-3}},\,10^{-4}\}$ \\
    \texttt{$lr_\text{d}$ (delays)} & $\{\mathbf{10^{-1}},\,10^{-2},\,10^{-3}\}$ \\
    \texttt{scheduler (weights)} & \{\textbf{one-cycle}, cosine, none\} \\
    \texttt{scheduler (delays)} & \{one-cycle, \textbf{cosine}, none\} \\
    \texttt{dropout} & 0.3 \\
    \texttt{epochs} & 100 \\
    \texttt{batch size} & 512 \\
    \hline
    \end{tabular}
    \label{tab:dcls_learning_parameters}
\end{table*}

\subsection{Hyperparameters Search}
\label{section:hyperparameters}

We detail here the hyperparameters that are searched for each of the three audio datasets for our non-delay C-SiLIF and SiLIF models. These grid searches are performed both for our C-SiLIF and SiLIF models and for all reproduced results on the AdLIF, cAdLIF, and RF models to ensure the most robust results in terms of training optimization. The only exception is for the AdLIF and cAdLIF results on GSC for Figure \ref{fig:sop} where configurations from corresponding papers are kept considering that the dataset and resolution are the same.

\begin{table}[h]
\centering
\caption{Hyperparameter grid search space per dataset for SiLIF and C-SiLIF.}
\begin{tabular}{@{}lccc@{}}
\toprule
\textbf{Hyperparameter} & \textbf{SHD} & \textbf{SSC} & \textbf{GSC} \\ \midrule
\texttt{lr} & $\{10^{-2}, 5 \cdot 10^{-3}, 10^{-3}\}$ & $\{10^{-2}, 5 \cdot 10^{-3}, 10^{-3}\}$ & $\{10^{-2}, 5 \cdot 10^{-3}, 10^{-3}\}$ \\
\texttt{dropout} & $\{0.1, 0.25, 0.5\}$ & $\{0.1, 0.25, 0.5\}$ & $\{0.1, 0.25, 0.5\}$ \\
\texttt{schedule\_patience} & $\{5, 10^*\}$ & $\{5, 10\}$ & $5$ \\
\texttt{schedule\_factor} & $\{0.7, 0.9^{*}\}$ & $0.7$ & $0.7$ \\
\texttt{C-SiLIF $\Delta t_{\text{max}}$} & $0.5$ & $5$ & $5$ \\ 
\texttt{batch size} & $128$ & $256$ & $32$ \\
\texttt{epochs} & $100$ & $100$ & $100$ \\
\bottomrule
\end{tabular}

\vspace{1em}
\begin{flushleft}
* Fixed for ablation study (Table~\ref{table:ablation}).
\end{flushleft}
\end{table}
The final obtained hyperparameters are presented in Table~\ref{tab:hyperparam}. All further fixed configurations are provided in our code base. 

\begin{table}[htbp!]
    \centering
    \caption{Experimental setups for our models on different datasets.}
    \begin{tabular}{lcccccc}
        \toprule
        \textbf{Dataset} & \textbf{Model} & \textbf{Resolution} & \textbf{\#Hidden Size} & \textbf{lr} & \textbf{Dropout} &  \textbf{\makecell{Scheduler\\(Patience/Factor)}} \\
        \midrule
        \multirow{2}{*}{SHD} & \multirow{2}{*}{C-SiLIF}
        & 4\,ms & 128 & $10^{-2}$ & 0.5 & 5/0.9 \\
        & & 4\,ms & 512 & $5 \cdot 10^{-3}$ & 0.1 & 10/0.9 \\
        \hline
        \multirow{2}{*}{SSC} 
        & C-SiLIF & 10\,ms/4\,ms  & 512 & $5 \cdot 10^{-3}$/$10^{-2}$ & 0.5 & 10/0.7 \\
        & SiLIF & 10\,ms/4\,ms & 512 & $10^{-2}$ & 0.5 & 10/0.7 \\
        \hline
        \multirow{2}{*}{GSC} 
        & C-SiLIF & 10\,ms & 512 & $10^{-3}$ & 0.25 & 5/0.7 \\
        & SiLIF & 10\,ms & 512/1024 & $10^{-3}$ & 0.5 & 5/0.7 \\
        \bottomrule
    \end{tabular}

    \label{tab:hyperparam}
\end{table}

\subsection{SHD SOTA Comparison}
\label{sec:shd}

We report in Table~\ref{table:results_shd} the extra SOTA comparison on the SHD task. On the SHD task, our C-SiLIF model outperforms previous state-of-the-art accuracy for networks below 50k parameters with 95.06$\%$ accuracy. At a larger scale, it nears the current state-of-the-art obtained for the Simplectic-Euler (SE)-adLIF model \cite{baronig2025}, within its margin of error. We note that this task is less relevant than the SSC and GSC datasets due to its limited number of samples and the absence of separate validation and test sets, leading to an over-fitted hyperparameter search. 

\begin{table}[htbp!]
\centering
\caption{Classification accuracy on SHD dataset.
Models are ranked by Top-1 accuracy (highest at the bottom) within non-delay and delay-based categories.
Our models are shown in bold.}
\begin{tabular}{lcccccc}
\toprule
\textbf{Method} & \textbf{Resolution} & \textbf{\# Params} 
& \textbf{Rec.} & \textbf{Del.} & \textbf{Top1 Accuracy} \\
\midrule

S5-RF \cite{huber2025} & 4\,ms & 0.2M &  &  & 91.86\% \\
cAdLIF \cite{co-learning-delays} & 10\,ms & 38.7k &  &  & 94.19\% \\
SE-adLIF, 1 layer \cite{baronig2025} & 4\,ms & 37.5k &  &  & 94.59 $\pm$ 0.27\% \\
RadLIF \cite{Bittar2022} & 14\,ms & 3.9M & \checkmark &  & 94.62\% \\
\textbf{C-SiLIF, 128 hidden units (ours)} & \textbf{4\,ms} & \textbf{38.7k} &  &  & \textbf{95.06 $\pm$ 0.37\%} \\
\textbf{C-SiLIF, 512 hidden units (ours)} & \textbf{4\,ms} & \textbf{0.35M} &  &  & \textbf{95.55 $\pm$ 0.28\%} \\
SE-adLIF, 2 layer \cite{baronig2025} & 4\,ms & 0.45M &  &  & 95.81 $\pm$ 0.56\% \\
\cline{1-6}
d-cAdLIF \cite{co-learning-delays} & 10\,ms & 76k &  & \checkmark & 94.85 $\pm$ 0.64\% \\
DCLS-Delays \cite{hammouamri2024} & 10\,ms & 0.2M &  & \checkmark & 95.07 $\pm$ 0.24\% \\
\bottomrule
\end{tabular}
\label{table:results_shd}
\end{table}

\subsection{Sparsity Analysis}

In addition to the synaptic operations (SOPs) shown in Figure \ref{fig:sop}, we report the specific sparsity values for different SNN models in Table~\ref{table:sparsity} for reference. The sparsity corresponds to the average amount of non-spiking timesteps per neuron produced with respect to the amount of total timesteps per sample. These high levels of sparsity on all models lead to a significant reduction of SOPs and contribute to making SNNs highly efficient for neuromorphic implementations.
\begin{table*}[ht]
\centering
\caption{Test sparsity at best validation epoch for main models on SHD and GSC datasets, all with 2 hidden layers of 512 units. Standard deviation is reported with $\pm$.}
\begin{tabular}{lcccc}
    \toprule
    \textbf{Dataset} & \textbf{AdLIF} & \textbf{cAdLIF} & C-\textbf{SiLIF} & \textbf{SiLIF} \\
    \midrule
    SHD, 4$ms$ resolution & 88.57 $\pm$ 0.56\% & 97.80 $\pm$ 0.13\% & 80.25 $\pm$ 0.33\% & 98.66 $\pm$ 0.09\% \\
    SSC, 4$ms$ resolution & N/A & 93.65 $\pm$ 0.11\% & 90.37 $\pm$ 0.28\% & 93.35 $\pm$ 0.20\%\\
    GSC, 10$ms$ resolution & 93.39 $\pm$ 0.65\% & 93.03 $\pm$ 0.02\% & 85.21 $\pm$ 0.47\% & 91.71 $\pm$ 0.20\%\\

    \bottomrule
\end{tabular}
\label{table:sparsity}
\end{table*}

\subsection{Event-SSM Synaptic Operations Computation}
\label{sec:ssm_ops}

We detail here the number of synaptic operations (SOPs) required per sample for the Event-SSM model \citep{schone2024}. This model achieves state-of-the-art performance on the SSC dataset but relies on projecting each event to a real-value vector and processing them without any sparsity, thus requiring a significant number of SOPs. We proceed through the SOPs computation for the smaller Event-SSM on SSC with 64 hidden states, as reported in Figure \ref{fig:sop}.

First, the Event-SSM relies entirely on event-based processing and lacks a sparsity mechanism, meaning each event triggers the same number of fixed operations throughout the system. The first operation each event goes through is an embedding layer, which projects it to a real-value vector of size 64. Then this vector goes through a first event-based SSM at its full resolution. The embedding is projected through the $\mathbf{\bar{B}}$ matrix, which is square here, meaning a cost of $64 \times 64$ SOPs/event. The state update is made through the diagonal matrix $\mathbf{\bar{\Lambda}}$, thus negligible in SOP, but the state is then projected further through the square matrix $\mathbf{\bar{C}}$ for again $64 \times 64$ SOPs/event. Considering the SSC dataset presents an average number of 8000 events per sample, as reported by \citet{schone2024}, the average SOP cost per sample for this first SSM block is already of $SOP_{\text{first block}} = 8000 \times (64 \times 64 + 64 \times 64) = 65.5M$. 

The model then uses a pooling operation that divides the number of events passing through the network by 8, resulting in 1000 events per sample in the second block. This block consists of two dense projections of $64 \times 64$ SOP/event and 3 more SSMs, for a total of $6 \times 64 \times 64$ SOP/event before the next event pooling. Altogether $SOP_{\text{second block}} = 1000 \times (2 \times 64 \times 64 + 6 \times 64 \times 64) = 32.8M$. The final pooling reduces the number of events to an average of 125 per sample in the final block, making it negligible in terms of SOP. 

Altogether, the 64 state Event-SSM model requires $SOP_{\text{total}} \leq SOP_{\text{first block}} + SOP_{\text{second block}}$ which corresponds to 98.3M SOPs/sample. One can simply adapt this computation to the 128-state SHD version of the Event-SSM model, leading to an even higher 288.8M SOPs/sample. 
These numbers are around 10 times higher than the measured values for our SNN models (Figure \ref{fig:sop}). Additionally, we note that the Event-SSM relies on full precision real-value number for all its activations, meaning that each synaptic operation requires a full precision multiply-and-accumulate (MAC) operation, whereas our binary spiking models can handle simple accumulate (AC) operations as synaptic values are either summed when there's a spike or else omitted. Overall, it appears that the Event-SSM is significantly more computationally intensive, especially on event-based neuromorphic hardware.

\subsection{SSM Models Optimization}
\label{sec:ssm_opti}

\subsubsection{S4D and SpikingSSM for Keyword Classification}
\label{sec:spikingssm}
We detail here the method to obtain results on the S4D \cite{S4D} and SpikingSSM \cite{shen2025spikingssms} presented in Figure \ref{fig:sop} in the main text. As mentioned above, \cite{shen2025spikingssms} already report results for these models. But these are obtained by processing the raw-audio data at a resolution of $0.06\,ms$ (vs $10\,ms$ to $4\,ms$ for our models). This leads to excessively high SOPs due to the repetition of operations at a very high frequency, leading to $3.1 \times 10^{9}$ SOPs for SpikingSSM and $5.7 \times 10^{9}$ SOPs for S4D, which is at least 1000 times more than our spiking models. Therefore, we do not report the results presented by \citet{shen2025spikingssms} and rather retrain and optimize these models in our standardized setup with a MFCC preprocessing at a resolution of $10\,ms$ for GSC. 

\subsubsection{Mamba for Keyword Classification}
\label{sec:mamba}
To these SSM models, we also add Mamba \cite{gu2024mamba}, a selective gated extension of S4D, which became a new standard for efficient large language models. Mamba modifies the SSM parameters $A$, $B$, and $C$ to be data-dependent such that writing or forgetting data for the state can be selective based on the input. This modification has opened an avenue for SSM-like models to tackle natural language processing tasks at scale, but the suitability of Mamba for shorter sequence edge tasks is still to be proven. \citet{ding2025keyword} recently proposed a bidirectional implementation of Mamba for keyword spotting, reaching high accuracies on the GSC dataset. \\
However, the use of bidirectional models prevents streamlined data processing, timestep by timestep, leading to a significant increase in computational costs. Indeed, the whole sequence needs to be buffered before it is sent to the bidirectional Mamba model, and the computation on the sequence needs to be repeated for each prediction, instead of only processing the latest timestep for standard forward models. This leads to 100 times more compute for GSC at $10\,ms$ resolution, doubled further due to the two paths of bidirectional models. The accuracy reported by \citet{ding2025keyword} is thus not comparable to our SNN models, and we retrain and optimize Mamba on GSC in a forward mode here.

\subsubsection{Experimental Setup}
To obtain the fairest results possible, as presented in Figure \ref{fig:sop} and \ref{fig:pareto_curve}, we explore the S4D, SpikingSSM, and Mamba models at different scales. We first re-use the model sizing proposed in \cite{shen2025spikingssms} with 6 hidden layers of model size 128 and state size of 32 for the SSM. This leads to SOPs between $2.0 \times 10^{7}$ and $4.0 \times 10^{7}$, which is around 10 times more than our models. Therefore, we also scale the models down to a medium size with 4 layers and 32 state size, and an extra small size for the S4D and Mamba with 3 layers of model size 64 and state size 32. For each of these scales, we conduct a grid search on the learning rate, dropout level, the $\Delta t_{max}$ initialization, and the normalization type between layernorm and batchnorm. We also maintain the special S4D optimizer setup with a scaled-down learning rate for the SSM parameters $A$, $B$, and $C$ (Equation \ref{eq:SSMeq}). The swept hyperparameters and sizes per model are reported in Table~\ref{table:ssm_hyper}, and the best obtained configurations for each model and scale are reported in Table~\ref{table:ssm_best}.

\begin{table}[h]
\centering
\caption{Hyperparameter grid search space for the SSM models on the GSC dataset.}
\begin{tabular}{@{}lccc@{}}
\toprule
\textbf{Hyperparameter} & \textbf{S4D} \cite{S4D} & \textbf{SpikingSSM} \cite{shen2025spikingssms} & \textbf{Mamba} \cite{gu2024mamba} \\ \midrule
\texttt{lr} & $\{5\cdot10^{-2}, 5 \cdot 10^{-3}, 10^{-3}, 5 \cdot 10^{-4}\}$ & $\{10^{-2}, 5 \cdot 10^{-3}, 10^{-3}\}$ & $\{10^{-2}, 5 \cdot 10^{-3}, 10^{-3}, 5 \cdot 10^{-4}\}$ \\
\texttt{dropout} & $\{0.1, 0.2\}$ & $\{0.1, 0.2\}$ & $\{0.1, 0.2\}$ \\
\texttt{normalization} & batch, layer & batch, layer & batch, layer \\
\texttt{$\Delta t_{\text{max}}$} & $\{0.1, 5.0\}$ & $\{0.1, 5.0\}$  & $\{0.1, 5.0\}$  \\ 
\texttt{layers} & $\{3, 4, 6\}$ & $\{4, 6\}$  & $\{3, 4, 6\}$ \\
\texttt{hidden units} & $\{64, 128, 128\}$ & $128$  & $\{64, 128, 128\}$ \\
\texttt{state size} & $\{32, 32, 64\}$ & $\{32, 64\}$  & $\{32, 32, 64\}$ \\
\texttt{batch size} & $32$ & $32$ & $32$ \\
\texttt{epochs} & $100$ & $100$ & $100$ \\
\bottomrule
\end{tabular}
\label{table:ssm_hyper}
\end{table}

\begin{table}[htbp!]
    \centering
    \caption{Obtained best hyperparameter configurations and scale for the SSM models on the GSC dataset.}
    \begin{tabular}{lccccc}
        \toprule
        \textbf{Model} 
        & \textbf{\makecell{Scale \\ (Layers / Hidden / State size)} }
        & \textbf{lr}
        & \textbf{Dropout} 
        & \textbf{Normalization} 
        & $\mathbf{\Delta t_{\max}}$\\
        \midrule
        
        \multirow{3}{*}{S4D \cite{S4D}}
        & 6 / 128 / 64 & $5\cdot10^{-3}$ & 0.2 & BatchNorm & 0.1 \\
        & 4 / 128 / 32 & $10^{-2}$ & 0.2 & BatchNorm & 5.0 \\
        & 3 / 64 / 32  & $10^{-2}$ & 0.1 & BatchNorm & 5.0 \\
        
        \midrule
        
        \multirow{2}{*}{SpikingSSM \cite{shen2025spikingssms}}
        & 6 / 128 / 64 & $5 \cdot 10^{-3}$ & 0.1 & BatchNorm & 5.0 \\
        & 4 / 128 / 32 & $5 \cdot 10^{-3}$ & 0.1 & BatchNorm & 5.0 \\
        
        \midrule
        
        \multirow{3}{*}{Mamba \cite{gu2024mamba}}
        & 6 / 128 / 64 & $10^{-3}$ & 0.2 & BatchNorm & 0.1 \\
        & 4 / 128 / 32 & $5 \cdot 10^{-3}$ & 0.2 & LayerNorm & 0.1 \\
        & 3 / 64 / 32  & $5 \cdot 10^{-3}$ & 0.2 & BatchNorm & 5.0 \\
        
        \bottomrule
    \end{tabular}
    \label{table:ssm_best}
\end{table}

\subsection{Incremental C-SiLIF Features Analysis}
\label{sec:incr_abl}

\begin{figure}[h]
\begin{center}
    \includegraphics[width=3.5in]{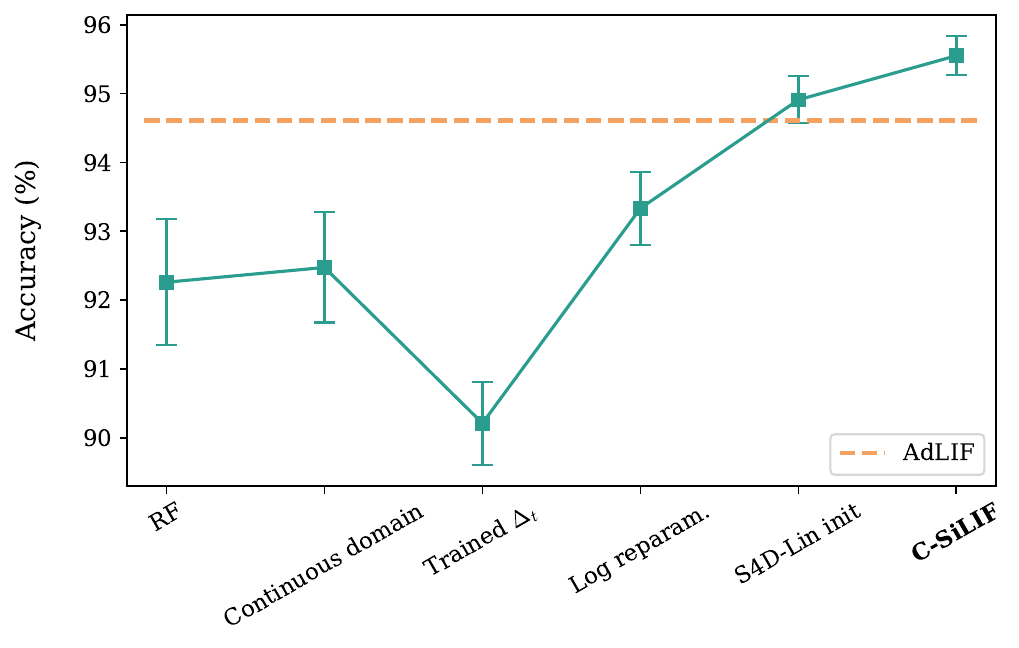}
\end{center}
\caption{Impact of incremental SSM-imported features on performance on the SHD dataset from the Resonate and Fire (RF) to our proposed C-SiLIF model.  Each model's performance is obtained after a grid hyperparameter sweep for 2 hidden layers of 512 neurons. Error bars correspond to the standard deviation on 5 seed runs.}
\label{fig:incr_study}
\end{figure}
We conduct an additional study of our C-SiLIF model, evaluating its performance as we incrementally add SSM-inspired features, from the Resonate and Fire (RF) neuron all the way to the C-SiLIF. We use networks of 2 hidden layers of 512 units and perform a grid search for each intermediate version. The measured average and standard deviation accuracy on the SHD dataset are reported in Figure \ref{fig:incr_study}. \\
As described by \citet{Izhikevich2001}, the RF neuron follows these neuronal dynamics:
\begin{equation}
\begin{aligned}
u_t & = u_{t-1} + \Delta t ((\alpha^{real} + i\alpha^{img})u_{t-1} + I_t ) - \theta s_{t-1} \\
s_t & = (Re(u_t) \geq \theta),
\end{aligned}
\label{eq:rf}
\end{equation}
where $\alpha^{real}$ and $\alpha^{img}$ are the trainable parameters, obtained directly from the discrete form.
As for the gap between the AdLIF and our SiLIF model, a first difference with the C-SiLIF is the parametrization of the model, as the C-SiLIF model focus on training continuous state transition parameters. The first incremental feature on top of the base RF model, called \textit{Continual domain} is thus to train $\lambda^{real}$ and $\lambda^{img}$ which are then converted back to the discrete transition variables $\alpha = exp((-\lambda^{real} + i\lambda^{img})\Delta_t)$ such that $\alpha^{real} = Re(\alpha)$ and $\alpha^{img} = Im(\alpha)$. We note that this step slightly improves the model's performance. 

The next step consists of making the timestep $\Delta_t$ a trainable parameter. Despite this feature having been identified as a central element of the success of our models (see ablation results in Table~\ref{table:ablation}), this incremental step hinders the model's performance. But as soon as the logarithmic reparametrization is included again, where $\lambda^{real}$ and $\Delta_t$ are trained in the logarithmic domain, the model outperforms the previous \textit{Continual domain} configuration. We thus hypothesize that these two features work in strong synergy and should be considered as one common method. \\
Further on, the model's parameter initialization is switched from the standard uniform distribution and range defined by \citet{higuchi2024} in their modern study of the RF model, to the \textit{S4D-Lin initialization} from the S4 model \citep{s4}. This leads to a significant jump in accuracy, reaching a level above our own optimized reproduction of the AdLIF model. This reinforces the idea that the specific S4 initialization has been optimized effectively for this specific model parametrization, and thus serves our C-SiLIF model. 

Finally, our proposed half reset and input gating $b$ (see Algorithm \ref{alg:dsiLIF_alg}) complete the C-SiLIF model and improve the final performance further. This all demonstrates the positive impact of the S4-inspired features along with their strong synergy for SNN models.

\end{document}